%% file: main.tex
\documentclass[envcountsame,runningheads]{llncs}

\usepackage{graphicx}                   %
\usepackage{hyperref}                   %
\usepackage{hieroglf}

\newtoggle{arxiv}
\toggletrue{arxiv}

\input{preamble/0_packages}
\input{preamble/0_macros}
\input{preamble/0_tikz}

\input{preamble/0_meta}
\AddToHook{cmd/appendix/before}{\crefalias{section}{appendix}}

\tikzset{
	state/.style={
		rectangle,
		rounded corners,
		draw=black,
		minimum height=2em,
		minimum width=2em,
		inner sep=4pt,
		text centered,
	}
}

\usepackage[Tol]{colorblind}
\usepackage{fontawesome5}
\usepackage[
    backgroundcolor=yellow]{todonotes}

\begin{document}

\pagestyle{plain}
\maketitle
\begin{abstract}
Synthesizing a reactive system from specifications given in linear temporal logic (LTL) is a classical problem, finding its applications in safety-critical systems design.
These systems are typically represented using either Mealy machines or AIGER circuits.
We present the second version of \SemML{}, which outperforms all state-of-the-art tools for finding either solution.
Aside from implementing the classical automata-theoretic approach, our tool utilizes partial exploration and machine-learning guidance for obtaining solutions efficiently, and numerous heuristics and improvements of classic algorithms for extracting small representations of these solutions.
We evaluate our tool against the existing state-of-the-art tools (in particular \Strix, \LtlSynt, and the previous version of \SemML) on the dataset of the synthesis competition SYNTCOMP.
We show that we solve significantly more instances and do so much faster than other tools, while maintaining state-of-the-art solution quality.
\end{abstract}

\input{chapters/01_introduction}

\input{chapters/02_tool_desc}
\input{chapters/03_semml_pipeline}

\input{chapters/06_experiments}

\section{Conclusion and Future Work}
With the implemented improvements at all stages and now the controller construction (synthesis in the narrower sense),
\SemML{} becomes the state-of-the-art tool for \emph{all} standard LTL synthesis tasks, namely realizability, Mealy machine synthesis, and AIGER circuit synthesis.
Methodologically, \SemML{} illustrates that machine learning can aid in safety-critical contexts without sacrificing correctness guarantees.
Practically, as \SemML{} constructs counterexamples efficiently, it becomes useful for counterexample-guided refinement approaches.

For future work, we identify several avenues.
First, it might be desirable to design an incremental variant of this synthesis pipeline.
Indeed, since queries in the CEGAR applications are often only slight derivations from each other, a lot of intermediate results can often be reused instead of recomputing them every time from scratch. 
Second, we believe that even the successor determinization can be integrated into the \texttt{MeMin} process, further increasing the potential for reductions.
Thirdly, we conjecture that there is still a lot of potential in reducing circuit size by finding appropriate state encodings, especially by further exploiting the semantic labelling.
Finally, another advantage to be explored is that controllers generated by \SemML{} are semantically labelled, which could provide e.g.\ runtime explanations or insight into counterexamples.

\paragraph{Data availability.} Our tool is continually developed at \url{https://gitlab.com/live-lab/software/semml}.
The artefact containing all tools, benchmarks, and scripts to reproduce our results is available at \url{https://doi.org/10.5281/zenodo.19763387}.

\bibliographystyle{splncs04}
\bibliography{main}
\newpage{}
\appendix

\iftoggle{arxiv}{
	\input{chapters/10_appendix}

}{
}

\end{document}

%% file: preamble/0_packages.tex
\usepackage{ellipsis, ragged2e} %
\usepackage[l2tabu, orthodox]{nag} %

\usepackage[english]{babel}
\usepackage[T1]{fontenc}
\usepackage[utf8]{inputenc}
\usepackage[final]{microtype}

\usepackage{bbm}
\usepackage{proof}
\usepackage{csquotes}
\usepackage{mathtools}
\usepackage{booktabs}
\usepackage{multirow}
\usepackage{afterpage}
\usepackage{xparse}
\usepackage{hyperref}
\usepackage{enumitem}
\usepackage[normalem]{ulem}
\usepackage{subcaption}
\usepackage{placeins}

\usepackage[table]{xcolor}

\usepackage[]{hhline}
\usepackage{nicematrix}

\usepackage[capitalise]{cleveref}

%% file: preamble/0_tikz.tex
\usepackage{tikz}

\usetikzlibrary{automata,positioning,shapes,calc,fit,arrows,arrows.meta,backgrounds,colorbrewer}
\usepackage{pgfplots}
\usepgfplotslibrary{fillbetween}
\pgfplotsset{compat=1.13}

\tikzstyle{state}+=[minimum size=20pt,inner sep=2pt]
\tikzstyle{action}=[font=\small,inner sep=0pt,outer sep=3pt]
\tikzstyle{actionedge}=[->,draw]
\tikzset{chainarrow/.tip={Stealth[length=3pt]}}
\tikzset{>=chainarrow}

\tikzstyle{system} = [draw,circle,minimum size=1cm]
\tikzstyle{environment} = [draw,diamond,minimum size=1cm]

\tikzstyle{systemltl} = [draw,rectangle,rounded corners=5pt,minimum size=1cm]
\tikzstyle{environmentltl} = [draw,rectangle,minimum size=1cm]

\tikzset{every picture/.append style={
	auto,
	node distance=2cm,
	initial text={},
}}

\pgfplotscreateplotcyclelist{no marks}{%
	loosely dotted, mark=\empty\\
	densely dotted, mark=\empty\\
	solid, mark=\empty\\
	densely dashed, mark=\empty\\
	loosely dashed, mark=\empty\\
	dashdotted, mark=\empty\\
	dashdotdotted, mark=\empty\\
	densely dashdotted, mark=\empty\\%
}

\newcommand{\plotmarksize}{1pt}
\newcommand{\axislines}[2]{
	\addplot[black,forget plot,update limits=false, no marks] coordinates {(0.001,0.001) (#1,#1)};
	\pgfmathsetmacro{\lowerCoordinate}{0.001*#2}
	\pgfmathsetmacro{\higherCoordinate}{#1/#2}
	\addplot[black,dashed,forget plot,update limits=false, no marks] coordinates {(0.001,\lowerCoordinate) (\higherCoordinate,#1)};
	\addplot[black,dashed,forget plot,update limits=false, no marks] coordinates {(\lowerCoordinate,0.001) (#1,\higherCoordinate)};
	
	\addplot[gray,thin,forget plot,update limits=false, no marks] coordinates {(#1,0.001) (#1,#1)};
	\addplot[gray,thin,forget plot,update limits=false, no marks] coordinates {(0.001,#1) (#1,#1)};
}

%% file: preamble/0_meta.tex
\title{
SemML 2.0: Synthesizing Controllers for LTL%
\thanks{%
This research has received funding from the European Union under Grant Agreement No.\ 101171844, ERC project Intelligence-Oriented Verification\&Controller Synthesis (InOVationCS).
Views and opinions expressed are, however, those of the authors only and do not necessarily reflect those of the European Union or European Research Executive Agency.
Neither the European Union nor the granting authority can be held responsible for them.
This research has also received funding from the MUNI Award in Science and Humanities MUNI/I/1757/2021 of the Grant Agency of Masaryk University.
}
}

\author{
	Jan~K\v{r}et{\'i}nsk{\'y}\inst{1,2} \orcidID{0000-0002-8122-2881} \textsuperscript{(\Letter)}
	\and Tobias~Meggendorfer\inst{3} \orcidID{0000-0002-1712-2165}
	\and Maximilian~Prokop\inst{1,2} \orcidID{0009-0008-6512-8693}
}

\institute{
	Masaryk University, Brno, Czech Republic
	\and
	Technical University of Munich, Germany
	\and
	Lancaster University Leipzig, Germany
}

%% file: chapters/01_introduction.tex
\section{Introduction}
Linear-Temporal-Logic (LTL) \cite{Pnueli77} reactive synthesis \cite{DBLP:conf/icalp/PnueliR89} is a prominent cornerstone in the long-standing quest of creating reliable software and hardware.
Intuitively, the task is to create a reactive system that adheres to a specification given in LTL, and thus is \enquote{correct by construction}. %
Additionally, LTL synthesis is a basic building block for synthesis problems in more complex domains.
For instance, LTL-modulo-theories synthesis \cite{RealizabilityModuloTheories,DBLP:conf/cav/RodriguezGS25} or synthesis over infinite arenas \cite{DBLP:conf/cav/AzzopardiSPS25} have recently emerged as extensions with practical motivation.
As LTL synthesis forms a subroutine in their counterexample-refinement loops, the demand for efficient and scalable LTL synthesis tools becomes even more prominent.

On the one hand, LTL synthesis is 2-EXPTIME-complete, and thus has long remained out of reach for practical applications.
On the other hand, recent advances in translating LTL formulae to small deterministic automata, e.g.~\cite{DBLP:journals/jacm/EsparzaKS20}, reignited practical interest in synthesis through the classical automata-theoretic approach \cite{vardi1986automata}. %
Indeed, the LTL tracks of the synthesis competition \Syntcomp{} \cite{DBLP:journals/corr/abs-2206-00251} have been won by tools based on these advances since 2018, e.g.~\Strix{} \cite{DBLP:conf/cav/MeyerSL18}. %

Furthermore, in contrast to the determinization of Safra \cite{DBLP:conf/focs/Safra88} and others \cite{DBLP:conf/lics/Piterman06,DBLP:conf/fossacs/Schewe09}, these translations additionally provide LTL \enquote{labels} for each automaton state, which intuitively describe its \emph{semantics} (i.e.\ what remains to be satisfied from here onward).
Exploiting this \emph{semantic labelling} through various means, including machine learning (ML), is the foundational idea of our tool \SemML{}.
In its previous version \cite{DBLP:conf/tacas/KretinskyMPZ25}, \SemML{} demonstrated how partial exploration of the automaton guided by ML evaluation of this labelling further increases the scalability of this approach.
In particular, \SemML{} won the \emph{realizability} track of \Syntcomp{}, i.e.\ determining whether a given specification is satisfiable or not.
However, additionally constructing a concrete system that does satisfy the specification (i.e.\ a witness to realizability) is highly relevant for most practical applications (and thus constitutes the \emph{synthesis} track).
Here, \SemML{} only implemented \enquote{textbook} approaches, without practical optimizations, locking away its significant potential for these use cases.

On the one hand, \Syntcomp{} only considers solutions in the form of AIGER circuits \cite{Biere-FMV-TR-07-1} (for fair comparison of different approaches), and only evaluates solutions for instances that are realizable in the first place (i.e.\ where a satisfying system exists).
Hence, many of the participating tools only focus on this specific application.
On the other hand, providing a reason why a given instance is not satisfiable (i.e.\ a counterexample) often is equally interesting, e.g.\ for the counter-example guided refinement loops in \cite{RealizabilityModuloTheories,DBLP:conf/cav/RodriguezGS25,DBLP:conf/cav/AzzopardiSPS25}.
Further, constructing the system (or counterexample) in the form of an explicit finite state machine (Mealy machine \cite{6771467}) is often desired or even required for several applications.
In particular, this new functionality of our tool has already been successfully applied to several of the mentioned refinement approaches (e.g.\ \texttt{sweap}\footnote{\scriptsize{ See~\url{https://github.com/shaunazzopardi/sweap/commit/83c1bc2cbfec9e0a40c8e0a81a4e233adfde780c}}} \cite{DBLP:conf/cav/AzzopardiSPS25}).
Altogether, aside from providing competitive circuits for realizable solutions, there is a need for more general forms of solution extraction.

\subsubsection{Our contributions.}
We address numerous practical shortcomings in the previous version of our tool, vastly improving the efficiency and quality of computed solutions.
To this end, we
\begin{itemize}[topsep=.5ex,itemsep=-.25ex,partopsep=.5ex,parsep=.5ex,label=$\bullet$]
	\item implement efficient extraction of both Mealy machines and AIGER circuits, both as witnesses and counterexamples,
	\item implement heuristical and SAT-based Mealy machine minimizations \dots
	\item \dots{} including a non-trivial extension of the \texttt{MeMin} algorithm \cite{Abel15},
	\item add various approaches towards circuit minimization, including several non-trivial extensions to our BDD library,
	\item significantly re-work the ML-guided exploration for further speed-ups, and
	\item add numerous engineering improvements to all stages of the pipeline.
\end{itemize}
We thoroughly evaluate the efficiency of our improvements on the entire benchmark set of \Syntcomp{} 2025, comparing to our previous version and other state-of-the-art tools \ltlsynt{} and \Strix{} (the previous winners of \Syntcomp{}).
We solve significantly more samples, and do so much faster, while providing solutions of comparable or even significantly better quality.
In particular, the Mealy machines produced by our tool are significantly smaller.

%% file: chapters/02_tool_desc.tex
\section{Tool Description}\label{sec:tool}
We provide an overview of our new version of \SemML{} by first formally stating the problem, describing how to use the tool, and explaining its high-level approach. 

\subsubsection{LTL synthesis.}
The problem of LTL reactive synthesis is defined as follows.
We are given an LTL \cite{Pnueli77} formula $\phi$ over a set of atomic propositions $\AP$ together with a partition of $\AP$ into \emph{environment} and \emph{system} propositions $\AP = \AP_{\smallenv} \union \AP_{\smallsys}$.
The environment and system generate an infinite word by repeatedly choosing which of \enquote{their} atomic propositions to enable in each step (with the environment typically choosing first). %
The system wins this \enquote{game} if the obtained word satisfies $\phi$.
The central question is whether the system has a \emph{winning strategy}, i.e.\ a way to choose which propositions to enable so that the combined word always satisfies the given formula.
In that case, the instance is called \emph{realizable} and \emph{unrealizable} otherwise (this decision problem is called \emph{LTL realizability}).
For example, the formula $\phi = \ltlGlobally (r \Leftrightarrow \ltlNext g)$ with $\AP_{\smallsys} = \{r\}$ and $\AP_{\smallenv} = \{g\}$ prescribes that whenever the environment sends a $r$equest, the system should in the next step $g$rant the request, and only then.
This formula is realizable, and a winning strategy is to remember whether the environment sent a request in the previous step.
The problem of \emph{LTL synthesis} asks for a concise representation of this winning strategy, typically as \emph{Mealy machine} \cite{6771467} or \emph{AIGER circuit} \cite{Biere-FMV-TR-07-1}, or report that no such strategy exists.

\subsubsection{Tool usage.}
\SemML{} supports explicit input via an LTL formula and a partition of the propositions or, equivalently, a TLSF file \cite{DBLP:journals/corr/Jacobs016} as used by SYNTCOMP.
It can solve both realizability and synthesis problems, where both output as Mealy machines or AIGER circuits is supported.
Further, \SemML{} additionally supports counterexamples in both formats (i.e.\ a winning strategy of the environment for unrealizable formulae). %
An example invocation looks like this:
\begin{verbatim}
	semml.py -f 'G(r <-> X g)' --ins=r --outs=g
\end{verbatim}
For compatibility and simplicity, we provide \SemML{} in a Docker container paired with a Python wrapper script, directly including the third-party dependencies \texttt{syfco}~\cite{DBLP:journals/corr/Jacobs016} (TLSF conversion), \texttt{abc}~\cite{abc} (AIGER circuit simplifier), and \texttt{kissat}~\cite{kissat} (efficient SAT solver).
To validate our implementation, we also implemented our own model checker for Mealy machines and circuits on top of numerous assertions.

\subsubsection{High-level architecture.}
\SemML{} solves LTL Synthesis via the \emph{automaton-theoretic} approach \cite{vardi1986automata}.
Here, the formula is first translated into an equivalent automaton (typically a {parity} automaton, DPA).
This automaton serves as an arena for a graph game where the winning condition for the system is the automaton's acceptance.
Thus, a winning strategy in this game means the system can force an accepting word in the automaton and thus satisfy the original formula.
While in theory, these steps happen sequentially, in practice, \SemML{} interleaves automata construction and game solving, and employs ML heuristics to guide the construction (see \cite{DBLP:conf/tacas/KretinskyMPZ25} for full details).

For synthesis, \SemML{} then encodes the winning strategy as a Mealy machine or an AIGER circuit.
Mealy machines naturally arise from the automata-theoretic approach, as they directly correspond to the reachable part of the automaton.
We employ several minimization heuristics and extend the \texttt{MeMin} approach to more general Mealy machines.
AIGER circuits are then obtained by applying the \emph{Huffman} procedure for converting the Mealy machine to a circuit \cite{HUFFMAN}.
Here, we again employ several heuristics and minimization procedures. %
An overview of this architecture and the improvements over the previous version is given in \cref{fig:tool-overview}.

%% file: chapters/03_semml_pipeline.tex
\section{Controller Synthesis with \textsc{SemML}} 

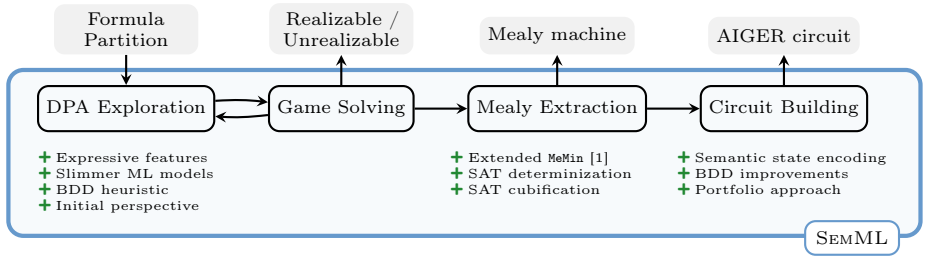
\begin{figure}[t]
	\centering
	\tikzset{
		partbox/.style={font=\scriptsize,minimum width=1.4cm,rectangle,rounded corners},
		toolstep/.style={partbox,minimum height=.6cm,draw,thick},
		iobox/.style={partbox,fill=gray!20,minimum height=.5cm},
		externalbox/.style={draw=gray,rectangle,rounded corners,text=darkgray!80,thick},
		stepedge/.style={->,>=stealth,thick},
		ioedge/.style={->,>=stealth,thick},
	}%
	\centering
	\newcommand{\addedicon}{\textcolor{green}{\faPlus}}
	\begin{tikzpicture}[node distance=0.5cm,remember picture]

	\node[toolstep] at (0,0) (explore) {DPA Exploration};
	\node[anchor=north,align=center,font=\tiny,text width=2.5cm] at (explore.south) (exploreexp)
	{
	\begin{itemize}[topsep=2pt,labelsep=2pt,label=\addedicon,leftmargin=9pt]
		\item Expressive features
		\item Slimmer ML models
		\item BDD heuristic
		\item Initial perspective
	\end{itemize}};
	
	\node[toolstep,right=.7cm of explore] (solve) {Game Solving};
	
	\node[toolstep,right=.7cm of solve] (mealy) {Mealy Extraction};
	\node[anchor=north,align=center,font=\tiny,text width=3cm] at (mealy.south) (mealyexp)
	{
	\begin{itemize}[topsep=2pt,labelsep=2pt,label=\addedicon,leftmargin=9pt]
		\item Extended \texttt{MeMin} \cite{Abel15}
		\item SAT determinization
		\item SAT cubification
	\end{itemize}};

	\node[toolstep,right=.7cm of mealy] (aiger) {Circuit Building};
	\node[anchor=north,align=center,font=\tiny,text width=3cm] at (aiger.south) (aigerexp)
	{
	\begin{itemize}[topsep=2pt,labelsep=2pt,label=\addedicon,leftmargin=9pt]
		\item Semantic state encoding
		\item BDD improvements
		\item Portfolio approach
	\end{itemize}};

	\node[iobox,above=.4 cm of explore,text width=1.7cm,align=center] (input) {Formula\\Partition};
	\node[iobox,above=.4 cm of solve,text width=1.7cm,align=center] (solveout) {Realizable / \\Unrealizable};
	\node[iobox,above=.4 cm of mealy] (mealyout) {Mealy machine};
	\node[iobox,above=.4 cm of aiger] (aigerout) {AIGER circuit};

	\begin{scope}[on background layer]
	\node[fit=(explore) (exploreexp) (solve) (mealy) (mealyexp) (aiger) (aigerexp),inner sep=.2cm,draw=T-Q-MC4,rectangle,rounded corners=.2cm,ultra thick,fill=T-Q-MC4!5] (dtbox) {};
	\end{scope}
	
	\node[draw=T-Q-MC4,rectangle,rounded corners=4pt,thick,fill=white,anchor=east,inner sep=4pt] at ([xshift=-.3cm]dtbox.south east){\scriptsize\SemML};

	\path[stepedge]
		(explore) edge[bend left=5] (solve)
		(solve) edge[bend left=5] (explore)
		(solve) edge (mealy)
		(mealy) edge (aiger)
	;
	\path[ioedge]
		(input) edge (explore)
		(solve) edge (solveout)
		(mealy) edge (mealyout)
		(aiger) edge (aigerout)
	;
	\end{tikzpicture}%
	\caption{
		Structural overview of \SemML's high-level architecture.
		All items marked by \addedicon{} are novel or significant improvements over our prior version.
		Additionally, several minor engineering improvements have been added to all stages.
	}
	\label{fig:tool-overview}
\end{figure}

As outlined in \cref{fig:tool-overview}, this version of \SemML{} comes with several advancements compared to its predecessor.
Due to space constraints, we provide an intuitive walk-through of the entire pipeline and highlight significant improvements with an intuitive explanation.
Detailed, technical descriptions can be found in \cref{app:advancements}.

\subsection{The Reactive Synthesis Pipeline of SemML}\label{sec:pipeline}
\subsubsection{Semantic LTL-to-automaton translation.}
As its predecessor, \SemML{} uses \Owl{} \cite{DBLP:conf/atva/KretinskyMS18} to convert the LTL formula into an equivalent automaton.
Internally, \Owl{} uses \emph{binary decision diagrams} (BDDs) \cite{bryant1986graph}, whose performance notoriously depends on their \emph{variable ordering}.
\SemML{} employs an intuitive heuristic to obtain a practically performant variable ordering (see \cref{app:ordering} for details).

Furthermore, the translations of \Owl{} are \enquote{semantic}.
This means that states are identified by a structured collection of LTL formulae (so-called \emph{semantic labelling}, example in \cref{app:translation}) and successors are purely derived from this labelling.
Intuitively, the semantic labelling describes the remaining language a state represents together with additional information required to track acceptance.
This comes with two useful properties that we exploit.
First, as this labelling fully captures the entire \enquote{context} of a state, a careful implementation can construct the automaton locally and on demand, i.e.\ given an automaton state, we can compute its successors without further context, which is also implemented in \Owl{}.
Second, the labelling of any state gives us an idea of what the automaton after that state might look like without actually computing it.
These two properties enable a \emph{guided partial exploration} approach, which is the core idea of our tool.

We provide full details of our construction and the resulting semantic labelling in \cref{app:translation}, including so far unpublished optimizations already used in the prior version of \SemML{} as well as \Strix.

\subsubsection{Guiding partial exploration.}
The (parity) automaton can be of double exponential size.
However, when solving the game, in practice, often only a small part of the arena is actually reachable and thus necessary for deciding the winner.
Thus, \SemML{} uses a partial exploration approach where it alternates between constructing the automaton and probing whether the currently constructed part already suffices to find a solution.
Ideally, this approach explores the \enquote{relevant} parts (i.e.\ the ones that will be reachable under a winning strategy) of the automaton first.
Therefore, \SemML{} uses machine learning based on the semantic labelling (i.e.\ feature-embeddings thereof) to heuristically identify winning choices and guide the exploration.
For a detailed discussion of the exploration algorithm and the interplay with the game solver, we refer to our previous version \cite{DBLP:conf/tacas/KretinskyMPZ25}.

Clearly, the practical gain of this approach heavily depends both on (i)~how well these models predict winning choices, and (ii)~how much overhead computing the guidance introduces.
In this work, we improve both points.
First, we introduce new, expressive features of the semantic labelling, allowing for better guidance (see \cref{app:features}).
Second, enabled by our new features, we simplify our model architecture to reduce inference time (see \cref{app:architecture}).
Moreover, as exploration happens from one player's perspective at a time, we introduce a new ML model to heuristically identify which player should explore first (see \cref{app:initpersp}).

\subsubsection{Extracting solutions.}
After finding a winning strategy, we need to encode it into the desired format (Mealy machines or AIGER circuits).
While both steps are simple theoretically, numerous heuristics are required to obtain practically competitive results.
In its previous version, \SemML{} only implemented the \enquote{textbook} method, without any sophisticated heuristics.
We fully reworked the approach, implementing improvements at all stages.
This allows us to solve much larger instances and obtain highly competitive controllers.

\paragraph{Mealy extraction.} Our game solver provides us with \emph{non-deterministic} strategies, meaning that instead of fixing a single choice in each state, it can allow for a set of choices where any choice is correct.
As such, it represents a set of strategies, i.e.\ a set of Mealy machines.
Our goal is to find a small Mealy machine within that set, which we achieve through several heuristics.
Our primary, SAT-based route works as follows:
We first search for a small (successor-deterministic) Mealy machine in this set.
Through a non-trivial extension of \textsf{MeMin} \cite{Abel15}, we then reduce this Mealy machine (i.e.\ without a determinization of the outputs that \textsf{MeMin} requires, see \cref{sec:memin}).
Only after the minimization, we further determinize the machine (see \cref{app:cubify}), which intuitively allows the minimization to consider a larger amount of different behaviors to find matching states or edges.
Since each of these steps involves numerous SAT queries, we added an efficient DIMACS interface to communicate with \texttt{kissat} \cite{kissat}.
Moreover, we employ several bounding heuristics and, in case the heuristical bound is too large, we switch to a faster but less efficient approach based on bisimulation reduction \cite{DBLP:conf/forte/RenkinSDP22}.

\paragraph{Circuit Extraction.} Obtaining circuits from Mealy machines then follows the Huffman procedure \cite{HUFFMAN}.
First, the states of the mealy machine are encoded as bit strings (known as the \emph{state encoding}), i.e.\ being in a particular state is represented by a particular combination of latches.
Then, the behaviour of every latch and output signal is translated to a circuit.

While any state encoding gives a correct circuit, finding an encoding that e.g.\ exploits certain symmetries in the Mealy machine can drastically reduce the size of the circuit.
Here, we provide an approach extending the (unpublished) \emph{structured} approach of \Strix, which is based on the semantic labelling of the Mealy machine (see \cref{app:encoding}).
(This also means that the semantic labelling needs to be preserved during the minimization step.)
To derive the circuit formulae, we follow the common approach to use binary decision diagrams (BDDs), as they naturally identify shared sub-formulae.
We implemented support for complement edges and \enquote{reduced-universe-simplification} in BDD library \texttt{JBDD} \cite{jbdd} for additional reductions (see \cref{app:bdd}).
Intuitively, a BDD with complement edges can represent $a \land b$ and $\neg (a \land b)$ using the same node (once complemented), whereas a classical BDD requires different nodes.
Further, when constructing the BDD nodes, we only need to ensure equivalence on actually reachable assignments.
This yields additional degrees of freedom that can be used to achieve smaller simplified formulae which are only equivalent on the \enquote{reduced universe} of reachable assignments, which can be achieved efficiently on BDDs.
(Formally, we can compute $g = \texttt{RESTRICT}(f, h)$, such that $g(x) = f(x)$ on all assignments $x$ where $h(x)$ is true, but $g$ potentially is much smaller than $f$, see e.g.\ \cite[Exercise~92]{knuth2011art}.)
As a last step, we employ the generic circuit reduction tool \texttt{abc} \cite{abc} to compress the final circuit even further.
As this pipeline contains many (partially optional) steps with complex interactions and it is often not clear which combination will yield the best results, we employ a portfolio of different configurations and return the smallest result (see \cref{app:portfolio} for details), similar to \Strix{}.

\subsection{Extended Mealy Minimization}\label{sec:memin}
Here, we provide insights into our approach towards minimizing non-deterministic Mealy machines in more detail, as it comprises a novel extension of the \texttt{MeMin} \cite{Abel15} algorithm, and can be used independently of our pipeline. 
Recall that the game solver provides us with a non-deterministic strategy.

\subsubsection{Successor determinism.}
A Mealy machine is \emph{successor deterministic} if for every state-input letter combination, the successor state of the machine is fixed (non-determinism in the output however is allowed).
To apply the \texttt{MeMin} approach, we first need to extract such a machine from the non-deterministic parity game solution.
Moreover, this machine should be as small as possible.
To this end, we first heuristically determine an upper bound (using a simple graph search).
If that upper bound is reasonably small, we then encode reachability and determinacy constraints into a SAT query and then search for a minimal model using counting constraints to reduce the strategy further.

\subsubsection{MeMin with non-deterministic outputs.}
Given such a (successor-deterministic) Mealy machine, the \texttt{MeMin} \cite{Abel15} approach essentially tries to find sets of states (\enquote{classes}) where all states of a class should for the same input be able to produce the same output and transition to (states of) the same successor class.
Then, a (potentially smaller) \emph{covering} Mealy machine is given by using the classes as new states, i.e.\ the new Mealy machine is consistent with the larger, non-deterministic one.
(Note that one state can belong to multiple classes.)

For a fixed number of classes, the relevant constraints can be cast into a SAT query using a clever encoding.
This allows us to search for the minimal required number of classes to obtain a covering machine.
Paired with several heuristics, this approach can achieve significant reductions.
However, to apply it to Mealy machines arising in our context, several adaptations are required (some of which are described in the original paper).

For example, instead of considering all $2^{\AP_{\sys}}$ different output symbols, we can derive \enquote{meta-symbols} representing sets of outputs.
In \cite{Abel15}, an approach for Mealy machines with \emph{cubes} as outputs is given.
(A cube effectively is given by a conjunction of (potentially negated) variables, e.g.\ $a \land b \land \lnot c$.)
Essentially, one derives a partitioning of $2^{\AP_{\sys}}$ such that every occurring output cube is a union of elements of this partition, and every such partition element becomes a meta-symbol.
While \cite{Abel15} assumes that machines have cube outputs (likely due to the used input file format), this is not the case for us.
To apply their approach, we thus would first need to \enquote{cubify} our output sets heuristically, eliminating potential for identifying shared outputs in the minimization.
While \Strix{} and \ltlsynt{} implement this approach, we adapt the core algorithm of \texttt{MeMin} to use arbitrary output sets.
This requires deep modifications, since the SAT encoding of \texttt{MeMin} is built around the assumption that if a set of states is pairwise compatible (i.e.\ each pair shares an output) then they are compatible as a whole (i.e.\ the entire set shares an output).
While this is implied by cube outputs, it no longer holds when we allow for arbitrary outputs.
We solve this by introducing additional constraints that ensure this consistency.
(Intuitively, we require that if state $x$ belongs to class $y$, then one of state $x$'s output partitions needs to be available in class $y$.)
At the same time, we aim to preserve the performance of the original approach and only introduce these constraints where necessary.

In terms of heuristics, \cite{Abel15} identifies an initial lower bound on the number of classes by heuristically identifying an as-large-as-possible set of pairwise incompatible states.
This problem effectively is a maximum clique search in the incompatibility graph.
As suggested in their paper, we also added a SAT-based maximization to further refine the heuristical lower bound.

%% file: chapters/06_experiments.tex
\section{Empirical Evaluation}
In this section, we aim to empirically answer the following research questions:
\begin{description}
	\item[RQ1] How efficient is \SemML{} at solving LTL synthesis tasks?
	\item[RQ2] How big is the impact of the improvements presented in this work?
	\item[RQ3] Do speed gains come at the cost of lower quality (i.e.\ larger) solutions?
\end{description}
To answer these questions, we evaluate our tool (and competitors) on the full \Syntcomp{} \cite{jacobs2024reactive} dataset (2025 iteration), comparing the number of solves, time required, and solution quality as defined in \Syntcomp{}.
We first describe the experimental setup, including benchmarks, metrics evaluated, and competing tools, before discussing the results for each question separately.
\subsection{Description}

\paragraph{Tools.}
We compare \SemMLTwo{} against the prior version \cite{DBLP:conf/tacas/KretinskyMPZ25} (written  \SemMLOne{} here), which won the Realizability track of \Syntcomp{} 2024 and came in second in 2025. %
Further, we compare to \Strix{} \cite{Strix} and \ltlsynt{} \cite{renkin2022dissecting}.
\Strix{} also follows a partial exploration approach and provides state-of-the-art solution quality, winning the LTL tracks in all iterations of \Syntcomp{} since its publication until stopping participation in 2023.
\ltlsynt{}, part of \spot{}, %
instead relies on highly optimized automata constructions and practical optimizations such as checks for immediately obvious strategies.
Further, \ltlsynt{} won the LTL tracks of \Syntcomp{} 2025.\footnote{Personal communication with the authors of \ltlsynt{} confirmed that it was not updated since 2024.
Rather, new benchmarks of \Syntcomp{} 2025 triggered a trivial-to-fix crash in SemML.}

\paragraph{Data.}
We use the \Syntcomp{} 2025 benchmark set, comprising 1585 LTL synthesis benchmarks.
Note that these are of widely varying difficulty:
\RealSolvedByAllInTenSec{} instances are solved by \emph{all} tools in under 10 seconds, and \RealSolvedByNoneInThirtyMin{} are solved by none in 30 minutes.
In other words, the set of \enquote{interesting} benchmarks is only a few hundred large, and being able to solve a few dozen more is significant.

We note that our ML models were trained on purely synthetic data, i.e.\ they have never \enquote{seen} \Syntcomp{} data.

\paragraph{Setup.}
We perform 3 separate runs with different targets over the entire benchmark set, namely (i)~deciding realizability (\runReal), (ii)~synthesizing a Mealy machine (\runMealy), and (iii)~synthesizing an AIGER circuit (\runAiger).
We ran each tool using the best-performing flags and options according to the last \Syntcomp{} they participated in.
Our experiments were run on a server with 40 cores of an Intel E5-2630 v4 CPU and 252GB of RAM.
As in \Syntcomp{}, we employ a timeout of 30 minutes per benchmark.

\paragraph{Metrics.}
We use the metrics of \Syntcomp{}.
Primarily, we are interested in the total number of instances solved within the timelimit.
Furthermore, on realizable formulae, solution quality is measured by a score, defined as follows.
Each solution receives $2-\log((s+1)/(r+1))$ points, where $s$ and $r$ are the size of the tool's solution and the size of the best known solution (i.e.\ the best of the 4 tools), respectively.
In other words, the best solution gets $2$ points, a solution that is one order of magnitude larger gets $1$ point.
This score then is averaged over all benchmarks a tool solved.
For Mealy machines, size is defined as the number of states and edges, and for AIGER circuits the number of latches and gates.
Note that the quality score is not directly penalized for timeouts, and thus should not be considered in isolation.

To compare runtimes, we compute the (geometric) mean %
of the ratios between \SemML's and other tools' runtimes per sample.
To avoid this ratio being blurred by constant time overhead (e.g.\ JVM startup and loading of ML model parameters),  %
we consider only non-trivial instances, i.e.\ where at least one tool required more than 30 seconds.
When a tool ran into the 30-minute timeout, we use that timeout value for the runtime ratio (benefiting tools with more timeouts). %

\subsection{Results and Discussion}
Our results are presented in \cref{tab:results}.
Based on these, we discuss each research question separately.
We provide an extensive analysis of our experiments, including detailed pairwise tool comparisons across all metrics, in the appendix.

\begin{table}[t]
	\centering
	\caption{Summary of our main experimental results.
		For every track, we report the number of solved instances and average solution quality where applicable (measured by the \Syntcomp{} sore, higher is better).
		Finally, we report the average speed-up of \SemML{} compared to other tools when constructing AIGER circuits (considering only non-trivial samples, where at least one tool required over 30 seconds, including timeouts).}
	\label{tab:results}
	\resizebox{0.9\textwidth}{!}{
		\begin{tabular}{l@{\hspace{15pt}}c@{\hspace{15pt}} c c@{\hspace{4pt}}c  c @{\hspace{15pt}}c@{\hspace{4pt}}cc @{\hspace{15pt}}c}
			&             \runReal             &  &                      \multicolumn{2}{c}{\runMealy}                       &  &                     \multicolumn{2}{c}{\runAiger}                     &  & Avg Speedup                  \\
			\cmidrule(l{2pt}r{16pt}){2-2} \cmidrule(l{2pt}r{2pt}){4-5} \cmidrule(l{2pt}r{2pt}){7-8} \cmidrule(l{2pt}r{2pt}){10-10}
			Tool &                Solves                 &  & Solves                                &           Quality            &  & Solves                                &          Quality          &  & $(> 30s)$                     \\
			\midrule
			\SemMLTwo                                                                                                                     &    \textbf{\SemmlRealSolvesTotal}     &  & \textbf{\SemmlMealySolvesTotal}       &    \SemmlMealyScoresMean     &  & \textbf{\SemmlAigerSolvesTotal}       &   \SemmlAigerScoresMean   &  &      \\
			\SemMLOne                                                                                                                     &       \SemmlTacRealSolvesTotal        &  & $-$                                   &             $-$              &  & \SemmlTacAigerSolvesTotal             & \SemmlTacAigerScoresMean  &  & \SemmlSemmlTacAigerTimeWtoFacThirty  \\
			\Strix                                                                                                                        &         \StrixRealSolvesTotal         &  & \StrixMealySolvesTotal                &    \StrixMealyScoresMean     &  & \StrixAigerSolvesTotal                &   \StrixAigerScoresMean   &  & \SemmlStrixAigerTimeWtoFacThirty     \\
			\ltlsynt                                                                                                                      &        \LtlsyntRealSolvesTotal        &  & \LtlsyntMealySolvesTotal              &   \LtlsyntMealyScoresMean    &  & \LtlsyntAigerSolvesTotal              &  \LtlsyntAigerScoresMean  &  & \SemmlLtlsyntAigerTimeWtoFacThirty  \\
		\end{tabular}
	}
\end{table}

\subsubsection{RQ1.}
We observe that \SemMLTwo{} solves the most LTL synthesis benchmarks of \Syntcomp{} across all three tracks. %
In particular, on \runMealy{} and \runAiger{}, our tool solves significantly more instances (77 and 45, respectively) than the prior state-of-the-art tool \Strix{}.
Recall that due to the small set of \enquote{challenging but feasible} instances in SYNTCOMP, this marks a major advancement. %
For runtime, we observe that \SemMLTwo{} offers an average speed-up of at least 2.5x compared to other tools.
Interestingly, we can observe that both \Strix{} and \ltlsynt{} solve more Aiger than Mealy cases.
This is especially surprising for \Strix{}, which also employs the automaton-theoretic approach and constructs an intermediate Mealy machine.
However, its Aiger-pipeline contains optimizations and heuristics which are not enabled when constructing Mealy machines.
Further data and analysis can be found in \cref{app:extended_exp}.

\subsubsection{RQ2.}
To assess the impact our presented improvements, we compare to the previous version \SemMLOne{}.
For realizability, we observe 51 additional solves, which we attribute to our improved exploration guidance and engineering improvements in the exploration phase.
More importantly, prior to this work, \SemMLOne{} only implemented an unoptimized \enquote{textbook} approach to extracting circuits, and did not support Mealy machine output.
Our new pipeline managed 102 additional solves on \runAiger{}.
Notably, only with these improvements, \SemML{} managed to surpass the prior state-of-the-art \Strix{}.

\subsubsection{RQ3.}
Comparing our average quality score with other tools, in particular \Strix{}, shows that our performance improvements do not come at the cost of solution quality (recall that quality is only computed on instances where a tool does not time out and should not be compared in isolation).
To further investigate the performance-quality trade-off, we also implemented a \enquote{fast mode}, which does not employ SAT-based reductions but only heuristics and the bisimulation-based reduction.
Here, we manage even more solves (\SemmlFastAigerSolvesTotal{} on the \runAiger{} track), however now with a steep quality reduction (down to \SemmlFastAigerScoresMean).
Further, note that while the average score is very similar to \Strix{}, there is quite a bit of variance among benchmarks.
In particular, there are samples where we produce much smaller circuits than \Strix{} and vice versa, hinting at great potential for a solver portfolio like \Neurosynt{} \cite{DBLP:conf/tacas/CoslerHOS24}.
For further details, see \cref{app:extended_exp}.

For Mealy machines, our solution quality even significantly exceeds that of \Strix{}, scoring a near-perfect 1.9. %
We conjecture two main reasons.
First, our extension of \textsf{MeMin} finds machines using fewer states.
This is due to more matching behaviors still being available, as we did not eliminate them through any upfront cubification.
Second, both \Strix{} and \ltlsynt{} seem to assume that the inputs of any edge are given as cubes, whereas we support edges with arbitrary Boolean expressions, heavily reducing the total number of edges required.
Again, for a more detailed analysis of this, we refer to \cref{app:extended_exp}.

\subsubsection{Summary.}
\SemMLTwo{} solves the most benchmarks across all three tracks and requires the least time to do so.
This is achieved without a notable decrease in the quality of AIGER circuits and furthermore accompanied by a notable increase in the quality of Mealy-machines (which in several cases are more relevant than AIGER circuits).
Altogether, based on our observations, \SemMLTwo{} is the best tool available for any LTL synthesis queries.

%% file: chapters/10_appendix.tex
\input{chapters/10_app_semantic_labelling}
\FloatBarrier

\input{chapters/10_app_further_advancements}
\FloatBarrier

\input{chapters/10_app_extended_experiments}

\FloatBarrier

%% file: chapters/10_app_semantic_labelling.tex
\section{LTL, Semantic Translation, and Semantic Labelling}
\newcommand{\fromIdx}[1]{\ensuremath{#1\text{:}}}
\subsection{LTL}
Linear temporal logic (LTL) is a logic over discrete time introduced in \cite{Pnueli77}.
Given a set of \emph{atomic propositions ($AP$)}, it allows for the qualitative description of infinite \textit{words} (or $\omega$-words) over the \emph{alphabet} $\Sigma = 2^{AP}$.
Intuitively, each \emph{letter} of a word denotes an assignment of $AP$ in which all atomic propositions contained in the letter are set to true whereas the others are set to false.
Formulae of LTL then contain statements about the development of these assignments over time.

Syntactically, an LTL formula in \emph{negation normal form (NNF)}, is given by the following grammar:
\begin{equation*}
	\varphi ::= \true\mid a\mid\neg \varphi\mid\varphi\land\varphi\mid \ltlNext \varphi\mid\varphi\ltlUntil\varphi
\end{equation*}
where $a\in AP$ and NNF refers to the fact that negations are only allowed in front of atomic propositions.

Furthermore, the semantics of an LTL formula is given by the satisfaction relation $\models$.
Let $w\in\Sigma^\omega$ be an $\omega$-word for which we define $w[i]$ to be its $i$-th letter and $w[\fromIdx{i}]$ the infinite suffix starting from letter $i$.
We say $w$ \emph{satisfies} a formula $\varphi$ ($w\models\varphi$) if it fulfills the following inductive relation:

\begin{align*}
	&w  \models \true&&  \\
	&w \models a && \text{iff.\ }a \in w[0]   \\
	&w \models \neg \varphi && \text{iff.\ }\neg(w \models \varphi) \\
	&w \models \varphi_1 \land \varphi_2 && \text{iff.\ } w\models \varphi_1 \text{ and } w \models \varphi_2\\ 
	&w \models \ltlNext \varphi && \text{iff.\ } w[\fromIdx{1}] \models \varphi \\
	&w \models \varphi_1\ltlUntil \varphi_2 && \text{iff.\ } \exists k.~ w[\fromIdx{k}] \models \varphi_2 \text{ and } \forall j<k.~ w[\fromIdx{j}] \models \varphi_1\\
\end{align*}

Based on the satisfaction relation, we define $\mathcal{L}(\varphi)=\{w \in (2^{AP})^{\omega}\mid w \models \varphi\}$ as the \emph{language} of $\varphi$, i.e.\ all the words that satisfy $\varphi$.

\subsection{Details on our DPA Translation and our semantic labelling}\label{app:translation}
In this section we describe our automaton construction in detail and discuss the resulting semantic labelling.
When translating LTL formulae to a parity automata, we use the modern route via Emerson Lei automata.
In particular, the procedure consists of the following 3 major steps:
\begin{enumerate}
	\item Normalize the LTL formula using the Normalization procedure of \cite{LTLNormalizationEsparzaSickert}.
	The result is a boolean combination of \enquote{simple-to-translate} subformulae. 
	\item Translate each individual subformula using the simple translations presented in \cite{DBLP:journals/jacm/EsparzaKS20}.
	The result is an Emerson Lei automaton.
	\item Translate the Emerson Lei automaton into a Parity automaton using the Zielonka tree construction \cite{ZielonkaTreeConstruction}.
	Here we use so-called \emph{conditional} Zielonka trees, which are an unpublished optimization implemented in \Owl{} and were used already in the previous version of our tool, as well as its predecessor \Strix.
\end{enumerate}
We proceed to talk about each of these steps in more detail.

\subsubsection{Normalization.}
From the Manna-Pnuelli Hierarchy \cite{Manna-Pnuelli-Hierarchy} we know that every LTL formula can be expressed as a positive boolean combination of Büchi and co-Büchi formulae (or an arbitrary boolean combination of Büchi formulae).
Note that in this definition, Büchi formulae include safety and co-safety formulae.
In \cite{LTLNormalizationEsparzaSickert} the authors finally prove this constructively, i.e.\ they show how to convert any arbitrary LTL formula into such a boolean combination of Büchi formulae.
This normalization procedure marks the first step in our automaton construction.

\subsubsection{Simple Translations for Fragments.}
In \cite{DBLP:journals/jacm/EsparzaKS20} the authors show how to easily convert Büchi formulae to Deterministic Büchi automata by using formula progression (also called \enquote{after function} or derivation).
Recall, that formula progression first uses unfolding equivalences (e.g.\ $\ltlFinally a \equiv a \lor \ltlNext\ltlFinally a$
to divide a formula into a \enquote{now}- and a \enquote{later}-part.
Then, by replacing the \enquote{now} part with a concrete assignment of propositions and checking which \enquote{later} parts remain, it is able to observe the effect of said assignment on the entire formula.
For example setting $a\mapsto\true$ in $a \lor \ltlNext\ltlFinally a$ yields $\true$ denoting no later goals remain (i.e. satisfaction of the formula) whereas $a\mapsto\false$ yields $\ltlFinally a$ denoting that the overall goal has not changed.

Using formula progression we can construct deterministic Büchi automata for safety and co safety by simply progressing the initial formula and declaring the sink states where the formula progressed to $\true$ as winning and symmetrically $\false$ as losing (i.e.\ everything else as winning).
States of these automata are identified by exactly one formula, the so called master formula.

This alone however, only suffices for pure safety or co-safety formulae, as only they will progress to these sinks.
To capture a formula like $\ltlGlobally\ltlFinally a$, we need to track the formula inside the $\ltlGlobally$.
To ensure this inner formula is satisfied infinitely often, we emit a single Büchi signal when the formula progresses to true and simultaneously reset the inner formula.
That way, the only way we can see infinitely many Büchi signals is by progressing the inner formula to true infinitely often, which exactly meany it is satisfied infinitely often.
We call this inner formula the \enquote{breakpoint formula} and together with the master formula, these two uniquely identify a DBA state.

Applying this procedure to all Büchi subformulae of the normalized formula yields a boolean combination of deterministic Büchi automata (which crucially includes negated deterministic Büchi automata or equivalently co-Büchi automata).
Identifying each DBA with a unique color, taking a full synchronous product over all of the DBAs and emitting a color whenever its corresponding DBA emitted a Büchi signal yields a deterministic emerson Lei automaton.
The acceptance condition is exactly given by the boolean combination of DBAs.

\subsubsection{Conditional Zielonka Trees.}
While in principle one could solve a graph game on a deterministic emerson Lei automaton \cite{DBLP:conf/fossacs/HausmannLP24}, this introduces practical difficulties, as such games require memory.
We prefer separating concerns and first introduce the required memory into the automaton (thereby converting it to a party automaton) and later solve a parity game, for which efficient algorithms are availible.
The transformation from Emerson Lei automata to parity automata uses the Zielonka tree construction as presented in \cite{ZielonkaTreeConstruction}.
In essence, this constructs a second automaton, which contains exactly the memory required for converting the Emerson Lei acceptance condition to a parity condition.
Intuitively, this automaton, observes the Büchi signals emitted by all the subautomata and emits \enquote{good} and \enquote{bad} parity colors depending on whether the observed combination of signal would be satisfying or rejecting, if seen infinitely often.
Thereby, seeing good combinations of subautomata signals infinitely often directly coincides with seeing good parity colors infinitely often.
Ultimately, the procedure is very comparable to the \enquote{Latest Appearance Record} presented in \cite{DBLP:conf/tacas/EsparzaKRS17} or the \enquote{Index appearance Record} from \cite{DBLP:journals/fmsd/EsparzaKS16}.
The parity automaton is then given by the product of the Emerson lei automaton and this second \emph{acceptance} automaton.

In practice, we can further optimize this conversion when we consider the semantic labelling.
Consider for example, a conjunction of automata, where one automaton tracks a transient property (like $\ltlFinally a$).
Since the overall formula will not be satisfied whithout this automaton seeing an $a$, there is no point in tracking the acceptance of any other automata at that point in time (and thereby possibly introducing extra states).
Only when an $a$ is observed, and the acceptance, is no longer \enquote{blocked} by a transient automaton, it makes sense to resume the acceptance tacking.
Similar reasoning applies if we can detect that an automaton is \emph{weak}, i.e.\ all of its strongly connected components are either fully accepting or fully rejecting.
This idea is known as \emph{conditional Zielonka trees} and is implemented in \Owl, and thus used in \Strix, and the previous version of \SemML.
However, it was never formally published.

\subsubsection{Full Example of our Semantic Labelling.}
\newcommand*\circled[2]{\tikz[baseline=(char.base)]{
		\node[shape=circle,draw,inner sep=1pt,fill=#2] (char) {\tiny #1};}}

\newcommand{\AoneAcc}{\raisebox{0.5mm}{\circled{1}{red!40}}}
\newcommand{\AtwoAcc}{\raisebox{0.5mm}{\circled{2}{blue!40}}}
\newcommand{\AthreeAcc}{\raisebox{0.5mm}{\circled{3}{green!40}}}
We present a full example for translating the formula $\varphi = \ltlFinally a \land \ltlGlobally\ltlFinally b \land \ltlFinally\ltlGlobally c$ to a DPA using this construction.
First, normalizing to a boolean combination of Büchi formulae yields $\varphi = \ltlFinally a \land \ltlGlobally\ltlFinally b \land \neg \ltlGlobally\ltlFinally \neg c$, in particular, the third subformula was changed.

Converting each individual Büchi subformula to a DBA yields the three automata in \cref{fig:subauts}.
Since $\ltlFinally a$ is a co-safety formula, no breakpoint formula is needed.
Further, observe, that for $\ltlGlobally\ltlFinally b$ and $\ltlGlobally\ltlFinally \neg c$ the overall obligation never changes and so neither does the master formula.
Additionally, the breakpoint formula is either satisfied immediately or remains unchanged.
Thus, these automata, only have one state with two self loops, depending on whether the letter satisfies the breakpoint formula or not.
Taking the synchronous product of all three DBAs yields the DELA in \cref{fig:dela} where the acceptance condition is the original boolean combination of subautomata.

To convert the DELA to a DPA, we construct a Zielonka Tree and subsequently a Zielonka automaton for the DELA's acceptance condition.
Observe that the acceptance condition of this DELA is blocked by the transient automaton for $\ltlFinally a$.
In other words, while we have not seen an $a$ we do not need to track the acceptance condition at all.
Further, after seeing an $a$, we no longer need to track the acceptance of $A_1$, as we know it progressed to true. 
Observe, that this information is available solely by investigating the semantic labelling of the subautomata.
This allows us to simplify the Acceptance condition to $\textit{inf}(\AtwoAcc) \land \neg\textit{inf}(\AthreeAcc)$ and only build a Zielonka automaton for that.
The resulting Zielonka automaton is given in \cref{fig:vonditionalZT}.
For comparisson, we also depict the regular Zielonka automaton in \cref{fig:regularZT}.
Observe, that without the semantic analysis, our Zielonka automaton has an extra state, and thus the resulting product DPA would be respectively larger.
Finally, we depict the resulting DPA in \cref{fig:dpa}.
The labelling of this DPA is given by the labelling of the DELA (a boolean combination of tuples of LTL formulae), and the Zielonka leaf.

\tikzstyle{aut state}=[draw, circle, inner sep=0.1cm]
\tikzstyle{sem aut state}=[draw, rectangle, rounded corners=0.2cm, inner sep=0.1cm]
\begin{figure}
	\begin{subfigure}{\textwidth}
		\centering
		\begin{tikzpicture}
			\node[sem aut state, initial] (q0) at (0,1.5) {\begin{tabular}{ll}
					M: & $\ltlFinally a$\\
					\midrule
					B: & $-$
			\end{tabular}};
			\node[sem aut state] (q1) at (2,1.5) {\begin{tabular}{ll}
					M: & $\true$\\
					\midrule
					B: & $-$
			\end{tabular}};
			
			\path[->]
			(q0) edge[loop above] node[right, near end] {$\neg a$} ()
			(q0) edge node[above] {$a$} (q1)
			(q1) edge[loop above, double] node[right, near end] {$\true$} ()
			;
			\phantom{\path[->]
				(q0) edge[loop below] node[right, near start] {$c$} ();}
			\node at (1,-0.2) {$A_1:\AoneAcc$};
		\end{tikzpicture}
		\begin{tikzpicture}
			\node[sem aut state, initial] (q0) at (0,1.5) {\begin{tabular}{lr}
					M: & $\ltlGlobally\ltlFinally b$\\
					\midrule
					B: & $\ltlFinally b$
			\end{tabular}};
			
			\path[->]
			(q0) edge[loop above, double] node[right, near end] {$b$} ()
			(q0) edge[loop below] node[right, near start] {$\neg b$} ()
			;
			\node at (0,-0.2) {$A_2:\AtwoAcc$};
		\end{tikzpicture}
		\begin{tikzpicture}
			\node[sem aut state, initial] (q0) at (0,1.5) {\begin{tabular}{lr}
					M: & $\ltlGlobally\ltlFinally \neg c$\\
					\midrule
					B: & $\ltlFinally \neg c$
			\end{tabular}};
			
			\path[->]
			(q0) edge[loop above, double] node[right, near end] {$\neg c$} ()
			(q0) edge[loop below] node[right, near start] {$c$} ()
			;
			\node at (0,-0.2) {$A_3:\AthreeAcc$};
		\end{tikzpicture}
	\caption{DBA for the simple formulae} \label{fig:subauts}
	\end{subfigure}
	
	\begin{subfigure}{\textwidth}
		\centering
		\begin{tikzpicture}
			\node[sem aut state] (q00) at (0,1.5) {\begin{tabular}{l}
					$\ltlFinally a$\\
					\midrule
					$-$
			\end{tabular}};
			\node[sem aut state,right= 1mm of q00] (q01) {\begin{tabular}{r}
					$\ltlGlobally\ltlFinally b$\\
					\midrule
					$\ltlFinally b$
			\end{tabular}};
			\node[sem aut state,right= 1mm of q01] (q02) {\begin{tabular}{r}
					$\ltlGlobally\ltlFinally \neg c$\\
					\midrule
					$\ltlFinally \neg c$
			\end{tabular}};

			\node (sf0) at ($ (q00.west)!0.5!(q02.east) + (0, .98cm) $) {$A_1\land A_2 \land \neg A_3$};
			\draw[] ($(q00.west)+ (0,.76)$) -- ($(q02.east)+ (0,.76)$);
			
			\node[sem aut state, initial,fit=(sf0)(q00)(q01)(q02)] (q0) {};
			
			\node[sem aut state] (q11) at (6.5,1.5) {\begin{tabular}{r}
					$\ltlGlobally\ltlFinally b$\\
					\midrule
					$\ltlFinally b$
			\end{tabular}};
			\node[sem aut state,right= 1mm of q11] (q12) {\begin{tabular}{r}
					$\ltlGlobally\ltlFinally \neg c$\\
					\midrule
					$\ltlFinally \neg c$
			\end{tabular}};
			\node (sf1) at ($ (q11.west)!0.5!(q12.east) + (0, 0.98cm) $) {$A_2 \land \neg A_3$};
			\node[sem aut state,fit=(sf1)(q11)(q12)] (q1) {};
			\draw[] ($(q11.west)+ (0,.76)$) -- ($(q12.east)+ (0,.76)$);
			
			\path[->]
			(q0) edge node[above] {$a$} (q1)
			
			(q0) edge[loop, in=110, out=130,  looseness=5] node[left=2mm,align=right] {$\neg a \land\neg b\land\neg c$\\$\AtwoAcc~\AthreeAcc$} (q0)
			(q0) edge[loop, in=50, out=70,  looseness=5] node[right=2mm,align=left] {$\neg a \land\neg b\land c$\\$\AtwoAcc$} (q0)
			(q0) edge[loop, in=310, out=290,  looseness=5] node[right=2mm,align=left] {$\neg a \land b\land c$\\$\phantom{\AoneAcc}$} (q0)
			(q0) edge[loop, in=250, out=230,  looseness=5] node[left=2mm,align=right] {$\neg a \land b\land\neg c$\\$\AthreeAcc$} (q0)

			(q1) edge[loop, in=110, out=130,  looseness=5] node[left=2mm,align=right] {$\neg b\land\neg c$\\$\AoneAcc~\AtwoAcc~\AthreeAcc$} (q1)
			(q1) edge[loop, in=50, out=70,  looseness=5] node[right=2mm,align=left] {$\neg b\land c$\\$\AoneAcc~\AtwoAcc$} (q1)
			(q1) edge[loop, in=310, out=290,  looseness=5] node[right=2mm,align=left] {$b\land c$\\$\AoneAcc$} (q1)
			(q1) edge[loop, in=250, out=230,  looseness=5] node[left=2mm,align=right] {$b\land\neg c$\\$\AoneAcc~\AthreeAcc$} (q1);
			;
		\end{tikzpicture}
	\caption{DELA $A_E=A_1\land A_2 \land \neg A_3$ with acc: $\textit{inf}(\AoneAcc) \land\textit{inf}(\AtwoAcc) \land \neg\textit{inf}(\AthreeAcc)$.
	}\label{fig:dela}
\end{subfigure}

\end{figure}
\begin{figure}
	\begin{subfigure}{\textwidth}
		\centering
			\begin{tikzpicture}
				\node[draw,rectangle,inner sep=2mm] (root) at (2,3) {$\AtwoAcc~\AthreeAcc$};
				\node[red] at (1.2,2.8) {1};
				\node[draw,rectangle,rounded corners=2.5mm,inner sep=2mm] (two) at (2,1.5) {$\AtwoAcc$};
				\node[red] at (1.4,1.3) {2};
				\node[draw,rectangle,inner sep=2mm] (empty) at (2,0) {$\emptyset$};
				\node[red] at (1.4,-0.2) {3};
				\node[] at (2.5,-0.2) {$l_0$};
				\path [-]
				(root) edge (two)
				(two) edge (empty)
				;
				
				\node[aut state] (q0) at (8,1.5) {$l_0$};
				
				\path[->]
				(q0) edge[loop, in=110, out=140,  looseness=10] node[left,align=right] {$\neg \AtwoAcc\land\neg \AthreeAcc$\\$\color{red}3$} (q0)
				(q0) edge[loop, in=345, out=15,  looseness=10] node[right,align=left] {$\AthreeAcc$\\$\color{red}1$} (q0)
				(q0) edge[loop, in=250, out=220,  looseness=10] node[left,align=right] {$\AtwoAcc\land\neg \AthreeAcc$\\$\color{red} 2$} (q0);
				;
				
			\end{tikzpicture}
		\caption{Conditional Zielonka Tree (left) for $\textit{inf}(\AtwoAcc) \land \neg\textit{inf}(\AthreeAcc$) with one leaf $l_0$ and corresponding Zielonka automaton (right) with one state.
		In the automaton the Büchi signals are the input letters and the red colors are the output parity colors.}
		\label{fig:vonditionalZT}
	\end{subfigure}
	
	\begin{subfigure}{\textwidth}
		\centering
		\begin{tikzpicture}
			\node[draw,rectangle,inner sep=2mm] (root) at (2,3) {$\AoneAcc~\AtwoAcc~\AthreeAcc$};
			\node[red] at (1,2.8) {1};
			\node[draw,rectangle,rounded corners=2.5mm,inner sep=2mm] (onetwo) at (2,1.5) {$\AoneAcc~\AtwoAcc$};
			\node[red] at (1.2,1.3) {2};
			\node[draw,rectangle,inner sep=2mm] (one) at (1,0) {$\AoneAcc$};
			\node[draw,rectangle,inner sep=2mm] (two) at (3,0) {$\AtwoAcc$};
			\node[red] at (0.4,-0.2) {3};
			\node[] at (1.6,-0.2) {$l_0$};
			\node[] at (3.6,-0.2) {$l_1$};
			\path [-]
			(root) edge (onetwo)
			(onetwo) edge (two)
			(onetwo) edge (one)
			;

			\node[aut state] (q0) at (6.5,1.5) {$l_0$};
			\node[aut state] (q1) at (9.5,1.5) {$l_1$};
			
			\path[->]
			(q0) edge[loop, in=165, out=195,  looseness=10] node[left,align=right] {$\neg \AtwoAcc\land\neg \AthreeAcc$\\$\color{red}3$} (q0)
			(q1) edge[loop, in=345, out=15,  looseness=10] node[right,align=left] {$\neg \AoneAcc\land\neg \AthreeAcc$\\$\color{red}3$} (q1)
			
			(q0) edge[bend left=100,  looseness=1.5] node[above,align=center] {$\AthreeAcc$ \\ $\color{red}1$} (q1)
			(q0) edge[bend left=20] node[above,align=center] {$\AtwoAcc \land \neg \AthreeAcc$ \\ $\color{red}2$} (q1)
			(q1) edge[bend left=20] node[below,align=center] {$\AoneAcc \land \neg \AthreeAcc$ \\ $\color{red}2$} (q0)
			(q1) edge[bend left=100,  looseness=1.5] node[below,align=center] {$\AthreeAcc$ \\ $\color{red}1$} (q0)
			;
			
		\end{tikzpicture}
	\caption{Regular Zielonka Tree for $\textit{inf}(\AoneAcc) \land\textit{inf}(\AtwoAcc) \land \neg\textit{inf}(\AthreeAcc$) with two leafs and corresponding Zielonka automaton (right) with two states.
		In the automaton the Büchi signals are the input letters and the red colors are the output parity colors.
		Observe how this automaton is more complex than the one for the conditional zielonka tree and that taking a product with this one doubles the state space of the Emerson lei automaton.}
	\label{fig:regularZT}
	\end{subfigure}
\end{figure}
\begin{figure}
		\centering
			\begin{tikzpicture}
				\node[aut state, initial] (q0) at (0,1.5) {$q_0,l_0$};
				\node[aut state] (q1) at (3,1.5) {$q_1,l_0$};
				
				\path[->]
				(q0) edge[loop above] node[right, near end] {$\neg a$} ()
				(q0) edge node[above] {$a$} (q1)
				(q1) edge[loop, in=110, out=140,  looseness=10] node[left,align=right] {$\neg b\land\neg c$\\$\color{red}1$} (q1)
				(q1) edge[loop, in=40, out=70,  looseness=10] node[right,align=left] {$\neg b\land c$\\$\color{red}2$} (q1)
				(q1) edge[loop, in=320, out=290,  looseness=10] node[right,align=left] {$b\land c$\\$\color{red}3$} (q1)
				(q1) edge[loop, in=250, out=220,  looseness=10] node[left,align=right] {$b\land\neg c$\\$\color{red} 1$} (q1);
				;
			\end{tikzpicture}
		\caption{DPA for the intersection of the DELA $A_E$ and the conditional Zielonka Tree.}
	\label{fig:dpa}
	
\end{figure}

%% file: chapters/10_app_further_advancements.tex
\section{Details on further advancements}\label{app:advancements}
In this section, we provide detail for all advancements mentioned in \cref{sec:pipeline}.
We describe then in order of appearance in \cref{sec:pipeline}.

\subsection{BDD Variable Ordering.}\label{app:ordering}
Internally, \Owl{} also uses BDDs to represent formulae, by creating one BDD variable per atomic proposition and sub-formula.
This allows for efficient (propositional) equivalence checks and numerous other operations.
One practically relevant choice for BDDs is the ordering of variables, which can have a drastic influence on size and runtime of BDD operations.
We get some constraints on the variable order from the automaton and game construction.
In particular, we need variables to appear before temporal subformulae, to efficiently compute the temporal step for all assignments at once.
Further, we need environment variables to appear before system variables to efficiently split the edge relation of the automaton and construct the game arena.
However, within the respective groups there are still degrees of freedom which we can use to optimize the variable order.

We propose to greedily order variables within a group based on how much fixing their value reduces the size of the formula.
For example, if we have $(a \land b) \lor (c \land \psi)$, where $\psi$ is some large formula, we prefer to decide $c$ first, as in one case, $\psi$ is removed entirely.
Thus, whatever variables are part of $\psi$ no longer need to be decided in the subtree where $c$ is false, and thus no longer contribute to the BDD size.
While seemingly trivial, this heuristic turned out to be quite helpful in practice and was a key factor for solving an entire family of formulae (the \texttt{mux} family).

\subsection{Control-Aware Features for Guidance}\label{app:features}
The driving idea of \SemML{} is to exploit the semantic labelling to identify promising exploration directions through machine learning.
Since our semantic labelling is rather complicated (a boolean combination of tuples of LTL formulae of arbitrary size), we employ machine learning.
This in turn requires a feature embedding of the semantic labelling, i.e.\ a fixed set of numeric values which capture key properties.
For example, one might consider how easy it is to satisfy a formula (measured by \emph{trueness} \cite{DBLP:conf/atva/KretinskyMM19}), and prefer directions which lead to higher trueness.

For many features, the \enquote{control} of atomic propositions / subformulae is highly relevant.
As an example, consider the formula $b \land \ltlNext a \lor \lnot b \land \ltlNext \Phi$, where $a$ is an environment variable and $b$ is a system variable.
Choosing $b$ would lead to the formula $a$, which in terms of raw trueness is ranked quite highly, however not a good choice since the trueness is not \enquote{controlled} by the system.
Previously, \SemML{} addressed this by computing a \enquote{controllability} feature, which was supposed to approximate this notion.
However, it was only used as an independent feature and fails to capture its key implications.
Thus, we integrate the notion of controllability into other features in two distinct ways.

\subsubsection{Sub-Automaton Attention.}
Recall that the semantic labelling of a automaton state effectively consists of a boolean combination of states of sub-automata, each of which effectively are labelled by one LTL formula (called \enquote{sub-formula} in the following).
Most of our features compute some concept (e.g.\ trueness) for every sub-formula and aggregate them using the boolean combination (see also \cite[section which]{DBLP:conf/tacas/KretinskyMPZ25}).
It turns out that quite often only very few sub-formulae are relevant for the current decision.
However, the \enquote{sharp} advice we can extract is then blurred by other sub-formulae during the aggregation (in particular, those controlled by the other player).
To address this issue, we first compute controllability (as in the previous version of \SemML) for every sub-formula.
Then, we remove the uncontrollable ones from the state formula by quantifying them out, assuming that the opponent resolves them in the worst possible way.
(Depending on the structure, this operation is either performed on the syntax tree or the BDD node.)
Finally, we then compute the aggregate value as in the previous version, however on the reduced formula.

\begin{example}
	Consider the formula $\varphi = \ltlGlobally\ltlFinally (b \leftrightarrow \ltlNext a) \rightarrow \ltlFinally\ltlGlobally(\neg b \land c)$, with $a$ being an environment variable and $b,c$ are system variables.
	It follows the common pattern of \enquote{assumptions-imply-guarantees} and is naturally decomposed into the boolean combination $\varphi = \neg\varphi_1 \lor \varphi_2$ with  $\varphi_1= \ltlGlobally\ltlFinally (b \leftrightarrow \ltlNext a)$ and $\varphi_2= \ltlFinally\ltlGlobally(\neg b \land c)$.
	Notice that $\varphi_1$ is controlled by the environment whereas $\varphi_2$ is controlled by the system.
	When computing a system feature, we would like $\varphi_1$ not to interfere with the feature value of $\varphi_2$ during aggregation.
	Thus, we eliminate $\varphi_1$ by assuming the environment falsifies it (as satisfying it would immediately satisfy the overall formula).
	Then, the remaining state formula is just $\varphi_2$, resulting in the desired sharp signal.
\end{example}
\subsubsection{MinMax with Lookahead.}
The concept of controllability is fundamentally ingrained into the graph game.
As such, by not just considering feature values in the direct successors, but rather using, e.g., a two-step lookahead and computing the min-max-value over it, we obtain a more faithful approximation, potentially at the cost of more exploration.
Comparing this min-max value to the current states value also gives an approximation of controllability which in turn can be used in the previous feature.
E.g.\ if the min-max-value of trueness goes down compared to the current state, this indicates, the environment has some controllability over the current state.
Note that in order not to create additional overhead, this feature only makes sense in situations where the edge relation of the subsequent states also available.
In particular, this feature is only used for the environment exploration heuristic, as when we split an automaton edge, we always get the edge relations for the newly created intermediate system states as well.

\subsection{Simpler Model Architecture}\label{app:architecture}
To enable faster inference, we improve our model architecture compared to \cite{DBLP:conf/tacas/KretinskyMPZ25}.
Recall that given a state and all of it's outgoing edges, our model essentially produces a ranking of all the outgoing edges, ordering them by believed likelihood to be the correct choice.
To do so we use \emph{pairwise ranking}, where we compare two edges at a time and say which of these we prefer, before aggregating all pairwise preferences into a single ranking.
While this produces rankings of high quality, it scales quadratically in the number of edges and thus introduces heavy inference overhead on larger instances.
Already in the prior version, we thus approximated this pairwise ranking by only sampling a subset of pairwise comparisons and approximate the ranking that way.
However, this was still flawed, as it prioritizes all parts of the ranking equally, whereas the by far most important part is to get the top choices correct.
In particular, it does not matter whether we correctly order the 102nd and the 103rd edge, as (ideally) nether of them is ever picked.

We thus introduce a new architecture that puts much more emphasis on getting the top edges correct.
First, we train a \emph{pointwise ranking} model, i.e.\ a model which assigns each edge a singular real number by which they can be ranked.
We use this to get an initial ranking and an idea of which edges should be ranked high.
To refine the top part of this ranking, we then resort to our strong pairwise ranking method.
In particular, we re-rank the top 8 elements of this ranking using a full round-robin tournament style to arrive at the final ranking of edges.
Note that this approximation is only employed when there are more than 16 edges to rank.
If there are fewer, we simply do a full round-robin tournament.

\subsection{Initial Exploration Perspective}\label{app:initpersp}
When exploring the game, we consider both player's perspectives (and alternate between them in a heuristical manner):
Intuitively, a player prefers a complicated region over one where they surely lose.
Thus, a good guidance for this player would lead us to the complicated region.
In contrast, the opponent's guidance would directly lead us to their \enquote{easy win}.
In this case, following the latter guidance first can identify a solution much quicker.
As such, we introduce another heuristic, again based on semantic labelling, to predict which player should go first.
Note that while this yields significant benefits for complicated samples, where such avoidable complicated regions exist, it introduces some additional initial overhead that can become significant for simple instances.

\subsection{Cubification of Mealy Machines}\label{app:cubify}
The classic BDD-based AIGER encoding approaches require Mealy machines with cubes as outputs, as this allows us to treat each output latch as individual formula / BDD (note that this usually \enquote{falls out} from the classic \texttt{MeMin} minimization approach).
Since we extended the minimization approach, our machines do not necessarily have cube outputs.
Hence, similar to successor determinization, we can choose which cubes to pick on each edge.
Intuitively, we want as many states as possible to share common outputs, since this should reduce the complexity of encoding the output relation.
Here, we first perform a greedy search for identifying an upper bound on the number of shared output cubes and then again apply a SAT-based search to refine this bound.

\subsection{Semantic State Encodings}\label{app:encoding}
For the conversion of mealy machines to circuits, we need a mapping from mealy states to bitstrings to encode the memory of the Mealy machine into combinations of latches of the circuit.
This is achieved by the so-called \emph{state encoding}.
Crucially, choosing the right state encoding can result in circuits orders of magnitudes smaller.

One desirable property of a state encoding is to exploit regularities or symmetries in the mealy machine.
Consider for example, a mealy machine whose states are divided in two disjoint sets $X,Y$ s.t. for every $x_i\in X$ there is a transition to some $y_i\in Y$ under the same proposition $a$.
This mealy machine is very regular or symmetric, as every state in $X$ has the same behavior under $a$.
A good state encoding could exploit this by dedicating a single bit to tracking in which set the machine currently is (e.g. 0 for $X$ and 1 for $Y$).
The value of this bit would easily be determined as we just need to check whether we still are in $X$ and whether the current letter is $a$ or not.
Thereby we can represent a lot of behavior of the mealy machine in a very small circuit.
While this Mealy machine sounds artificial, this is exactly the structure one would encounter for a formula of the form $\varphi \land \ltlFinally a$, where $\varphi$ does not contain $a$.
In particular $X$ and $Y$ would be copies of each other that each track $\varphi$ and for every state in $X$ (where $\ltlFinally a$ is not yet satisfied) there is a counterpart in $Y$ (where it \emph{is} satisfied) reachable via $a$.

Motivated by this example, we introduce a state encoding based on the semantic labelling and in particular on the subformulae, that appear or do not appear in each state.
Recall that the semantic labelling is given by a boolean combination of tuples of LTL formulae, as well as a leaf of the Zielonka Tree.
Intuitively, we encode each part of the semantic labelling separately, and concatenate the results.
For the governing boolean formula, we simply enumerate all different formulae and bitblast (use the binary representation of the number) their unique ID.
For LTL formulae, however, we first look at all the different subformulae that appear throughout the entire machine at that particular part of the semantic labelling.
E.g.\ we assemble all different master formulae of the first sub-automaton, across the entire machine.
Then, we create a bit for every subformula and encode the formula by setting each bit to 1, exactly if that subformula appears in that formula or not.
As this does not distinguish between $\ltlFinally a \land \ltlGlobally b$ and $\ltlFinally a \lor \ltlGlobally b$, we ad an additional deduplication part at the end.
However, since in practice the master formulae formulae are derivations of each other, this rarely happens.

We note that this is heavily inspired by \Strix{} which already introduced a similar encoding in \cite{DBLP:journals/acta/LuttenbergerMS20} and later updated it to the semantic labelling of our automaton.
However, where things become different is the compression of the encoding, we employ afterwards.
Very often this encoding is larger than it needs to be and contains a lot of redundant bits.
E.g.\ for a formula containing $\ltlGlobally\ltlFinally a$, every derivation will contain this subformula, as $\ltlGlobally\ltlFinally a$ is neither satisfied nor falsified by any finite prefix.
Thus, the bit representing this subformula will be $1$ in every single encoding.
Additionally the encoding might include more detail than necessary.
E.g.\ for a formula of the form $\ltlNext a \land \ltlNext b \land \varphi$, the initial state is the only state to ever contain the subformulae $\ltlNext a $ and $\ltlNext b$ (assuming they never reappear through $\varphi$).
Thus, it's encoding differs from every other states by two bits whereas one would suffice.
To tackle both of this, we propose the following encoding compression.
We greedily identify a locally maximal set of bits to remove s.t.\ after removal, all state encodings are still pairwise unique.
To efficiently compute this, we again use BDDs.
After that, we simply cut out these bits from the encoding, often obtaining a much smaller state encoding.

\subsection{Bdd Reductions}\label{app:bdd}
To assemble the circuit formulae for each latch and output signal, we again use BDDs as they naturally identify shared subformulae.
To further reduce the size of the BDD, we implement complement edges and \enquote{reduced universe simplification} (formally known as \texttt{RESTRICT} in \cite[Exercise~92]{knuth2011art})

The benefit of complement edges can be seen with the example from the main body.
Consider a circuit, which has two signals $o_1,o_2$ which are given by $o_1 = a \land b$, and $o_2 = \neg(a \land b)$, where $a$ and $b$ are input signals.
If spelled out like that, it is obvious that we should build the conjunction of $a$ and $b$ and wire its  output to $o_1$ and the negation of the output to $o_2$.
However, a classical BDD would simplify $o_2$ to $\neg a \lor \neg b$, yielding a different formula and thus yielding a larger circuit.
A BDD with complement edges, can recognize that $o_2$ is exactly the negation of $o_1$ and point to the same node, just once it is inverted.

Formally, the \texttt{RESTRICT} operation computes $g = \texttt{RESTRICT}(f, h)$, such that $g(x) = f(x)$ on all assignments $x$ where $h(x)$ is true, but $g$ potentially is much smaller than $f$, see e.g.\ \cite[Exercise~92]{knuth2011art}.
We can make use of this by defining $h$ to be the set of reachable assignments of the circuit which when using the semantic state encoding from above, could be only a small fraction of all possible assignments.
Consider for example a formula $\phi = a \land (b \lor c)$ and a set of reachable assignments $A = \{(0,1,0),(0,0,1),(1,1,0),(1,0,1)\}$ with order $(a,b,c)$.
The keen eye (or the BDD engine) might recognize that in all of these reachable assignments $(b \lor c) \equiv 1$.
Thus, if we want to simplify $\phi$ on this reduced universe of $A$ (i.e. $\texttt{RESTRICT}(\phi, A)$), we just obtain the simple formula $a$. 

\subsection{Portfolio}\label{app:portfolio}
Our circuit extraction pipeline contains many (partially optional) steps with complex interactions and it is often not clear which combination will yield the best results.
Thus, similar to \Strix{}, we employ a portfolio of different configurations and return the smallest result (see \cref{app:portfolio} for details).
In particular, we take the best result out of the following combinations:
\begin{enumerate}
	\item minimize the machine and use semantic encoding
	\item minimize the machine and use binary encoding
	\item do \textbf{not} minimize the machine and use semantic encoding
	\item do \textbf{not} minimize the machine and use binary encoding
\end{enumerate}
The reasoning is that as already noted by \Strix \cite{DBLP:journals/acta/LuttenbergerMS20}, the minimization can sometimes destroy the structure that the semantic encoding is trying to capture.
Thus, we also attempt the semantic encoding on the non minimized machine and the binary (smaller but unstructured) encoding on the minimized machine.
The last combination is there only for completeness and rarely is the one yielding the final circuit.

%% file: chapters/10_app_extended_experiments.tex
\section{Extended experiments: Pairwise tool comparison}\label{app:extended_exp}
In this section we provide an in-depth pairwise comparison between \SemMLTwo{} and the other tools we evaluate against, namely \SemMLOne, \Strix, and \ltlsynt.
We provide further material to investigate each metric individually.

\subsection{Number of instances solved} 
For all 3 Tracks, we provide pairwise confusion matrices about the number of solves (\cref{fig:confusion_matrices}).
Besides \SemML{} solving most benchmarks in all tracks, there are a few interesting additional takeaways here.

\subsubsection{Unique solves of \ltlsynt.} 
First, we look at the unique solves on the \runReal{} track.
While we mostly subsume our old version and \Strix{}, \LtlSynt{} has a notable amount of unique solves.
Almost all of these samples follow one of two shapes.
First, we are looking at samples (e.g.\ variants of arbiters) where the entire state space is necessary i.e.\ partial exploration cannot yield any benefit and solely introduces overhead.
Second, \LtlSynt{} checks for the existence of directly obvious strategy before starting the full synthesis procedure.
This allows them to e.g.\ solve the entire family \texttt{shift} family where specs are of the form $\ltlGlobally (e_i \leftrightarrow s_{i+1})$, where $e_i,s_i$ are environment and system variables respectively, and  $i<N$ for some $N$
\LtlSynt{} recognizes that there is a direct and unhindered correspondence between $e_i$ and $s_{i+1}$ on the formula level and does not even start to construct the astronomical automaton that arise when $N$ gets large.

When looking at the synthesis tracks, the number of unique solves is much smaller.
This is because while \ltlsynt{} can \emph{decide} the complicated arbiters whose entire state space is required, it still struggles to extract the solution.
For the samples, where a direct strategy was found, this is often much easier and thus, these samples mostly make up the unique solves of \ltlsynt{} on the \runAiger{} track.

\subsubsection{\Strix{} solves more Aiger than Mealy.} 
As a Mealy machine is an intermediate step to construct an Aiger circuit in \Strix, this seems counter intuitive at first glance.
However, when comparing the difference unique solves (and in fact the total solves from the main body as well) for \Strix{} on the \runMealy{} and \runAiger{} track, this seems to be the case.
This difference is caused by \Strix{} having a dedicated Aiger-portfolio that contains several empirical cutoffs to decide which subroutines to use to obtain small solutions and which ones to skip, because intermediate results are too large and it is unlikely that the procedure terminates within the customary timeouts.
As \Strix{} lacks this kind of portfolio for mealy extraction, we observe vastly different numbers here.

\subsection{Time}
For all 3 tracks, we provide pairwise scatterplots to compare the tools benchmark by benchmark (\cref{fig:time_scatter}).
We further provide Speedup factors for different lower cutoffs and comment on any additional insights (\cref{fig:speedup_factors}).

\subsubsection{Impact of \SemML's Solution Extraction Pipeline.} 
As already highlighted in the main body of this paper, the impact of our newly implemented solution pipeline is tremendous.
We can observe this by comparing the rather symmetrical (except for many timeouts) scatter plot for \SemMLOne{} on the \runReal{} track, and the very asymmetrical plot on the \runAiger{} track.
The vast majority of samples heavily profits from our new pipeline, despite additional steps like Mealy machine minimization being introduced.

\subsubsection{Constant time overhead for Startup.}
\SemML{} requires some time to start up and even start actually solving a sample, whereas the compiled tools \Strix, and \ltlsynt start solving the sample right away.
The main contributors to our delay at startup are (i) starting the JVM (ii) loading the parameters of our ML models and creating the inference objects, and (iii) setting up the infrastructure for our exploration algorithm and (iv) the python wrapper that e.g.\ read config files, and prepares the correct invocation of the Java-side.
For easy samples, this can heavily dominate the runtime and result in the phenomenon that can be observed in the lower left corner of the plots for \Strix{} and \ltlsynt{}.
Since the main goal of \SemML{} is to target samples of previously infeasible complexity, some seconds of constant overhead is very tolerable.
However, in order to focus on said complex samples, these small instances often need to be removed.

\subsubsection{Effect of Different Lower Cutoffs on Speedup Factor.}
To see the effect of various lower cutoffs, observe \cref{fig:speedup_factors}.
While for a cutoff of 0 (i.e.\ all samples) this effect is still very prominent, already a cutoff of 5 and 10 (on the \runAiger{} track) suffice to surpass \Strix{} and \LtlSynt{}, respectively.
Notice how this heavily reduces the number of considers instances, which highlights the nature of the \Syntcomp{} dataset of containing a lot of very easy instances.
However, it still leaves a healthy remainder of several hundreds for most cutoffs, and at least one hundred for the final two minute cutoff.

Further, observe how the speedup factor increases when we increase the focus on more and more complex instances.
This demonstrates that \SemML{} succeeds in its major target of bringing down the runtime of large instances by significant factors.
Ultimately, this further underpins the findings presented in the main body.

\subsection{Solution Quality.}
\begin{figure}[p]

	\pgfplotsset{
		table/col sep=comma,
		plot/.style={
			width=\textwidth,height=\textwidth,
			x label style={anchor=north,inner sep=0pt},
			y label style={anchor=south,inner sep=0pt},
			xtick={10,100,1000,10000}, xticklabels={10,100,1k,10k}, 
			ytick={10,100,1000,10000}, yticklabels={10,100,1k,10k},
			xmin=5,ymin=5,xmax=10000,ymax=10000,
			axis x line*=bottom,
			axis y line*=left
		},
		status colors/.style={
			scatter,
			scatter src=explicit symbolic, %
			scatter/classes={
				REALIZABLE={draw=Dark2-C, mark=x},
				UNREALIZABLE={draw=Dark2-C, mark=x}
			}
		}
	}
	
	\begin{subfigure}{\textwidth}
		\begin{minipage}{0.32\textwidth}
				\begin{tikzpicture}
					\begin{axis}[plot,status colors,xmode=log,ymode=log,xlabel=\SemMLOne,ylabel=\SemML]
						\axislines{10000}{10}
						\addplot+[only marks,mark size=\plotmarksize]
						table [x=size_SemmlTac, y=size_Semml, meta=status_gt] {data/helper_csvs/pairwise_mealy_size_Semml_SemmlTac.csv};
					\end{axis}
				\end{tikzpicture}
			
		\end{minipage}
		\begin{minipage}{0.32\textwidth}
			\begin{tikzpicture}
				\begin{axis}[plot,status colors,xmode=log,ymode=log,xlabel=\Strix]
					\axislines{10000}{10}
					\addplot+[only marks,mark size=\plotmarksize]
					table [x=size_Strix, y=size_Semml, meta=status_gt] {data/output_csvs/pairwise_mealy_size_Semml_Strix.csv};
				\end{axis}
			\end{tikzpicture}
		\end{minipage}
		\begin{minipage}{0.32\textwidth}
			\begin{tikzpicture}
				\begin{axis}[plot,status colors,xmode=log,ymode=log,xlabel=\ltlsynt]
					\axislines{10000}{10}
					\addplot+[only marks,mark size=\plotmarksize]
					table [x=size_Ltlsynt, y=size_Semml, meta=status_gt] {data/output_csvs/pairwise_mealy_size_Semml_Ltlsynt.csv};
				\end{axis}
			\end{tikzpicture}
		\end{minipage}
		\subcaption{Pairwise quality information of the \runMealy{} track, measured by number of states + number of edges.}
	\end{subfigure}
	
	\pgfplotsset{
		table/col sep=comma,
		plot/.style={
			width=\textwidth,height=\textwidth,
			x label style={anchor=north,inner sep=0pt},
			y label style={anchor=south,inner sep=0pt},
			xtick={1,10,100}, xticklabels={1,10,100}, 
			ytick={1,10,100}, yticklabels={1,10,100},
			xmin=1,ymin=1,xmax=100,ymax=100,
			axis x line*=bottom,
			axis y line*=left
		},
		status colors/.style={
			scatter,
			scatter src=explicit symbolic, %
			scatter/classes={
				REALIZABLE={draw=Dark2-C, mark=x},
				UNREALIZABLE={draw=Dark2-C, mark=x}
			}
		}
	}
	
	\begin{subfigure}{\textwidth}
		\begin{minipage}{0.32\textwidth}
			\begin{tikzpicture}
				\begin{axis}[plot,status colors,xmode=log,ymode=log,xlabel=\SemMLOne,ylabel=\SemML]
					\axislines{100}{10}
					\addplot+[only marks,mark size=\plotmarksize]
					table [x=size_SemmlTac, y=size_Semml, meta=status_gt] {data/helper_csvs/pairwise_mealy_size_Semml_SemmlTac.csv};
				\end{axis}
			\end{tikzpicture}
			
		\end{minipage}
		\begin{minipage}{0.32\textwidth}
			\begin{tikzpicture}
				\begin{axis}[plot,status colors,xmode=log,ymode=log,xlabel=\Strix]
					\axislines{100}{10}
					\addplot+[only marks,mark size=\plotmarksize]
					table [x=size_Strix, y=size_Semml, meta=status_gt] {data/output_csvs/pairwise_mealy_state_size_Semml_Strix.csv};
				\end{axis}
			\end{tikzpicture}
		\end{minipage}
		\begin{minipage}{0.32\textwidth}
			\begin{tikzpicture}
				\begin{axis}[plot,status colors,xmode=log,ymode=log,xlabel=\ltlsynt]
					\axislines{100}{10}
					\addplot+[only marks,mark size=\plotmarksize]
					table [x=size_Ltlsynt, y=size_Semml, meta=status_gt] {data/output_csvs/pairwise_mealy_state_size_Semml_Ltlsynt.csv};
				\end{axis}
			\end{tikzpicture}
		\end{minipage}
		\subcaption{Pairwise quality information of the \runMealy{} track, measured by number of states.}
		\label{subfig:states}
	\end{subfigure}
	
	\pgfplotsset{
		table/col sep=comma,
		plot/.style={
			width=\textwidth,height=\textwidth,
			x label style={anchor=north,inner sep=0pt},
			y label style={anchor=south,inner sep=0pt},
			xtick={10,100,1000,10000}, xticklabels={10,100,1k,10k}, 
			ytick={10,100,1000,10000}, yticklabels={10,100,1k,10k},
			xmin=1,ymin=1,xmax=10000,ymax=10000,
			axis x line*=bottom,
			axis y line*=left
		},
		status colors/.style={
			scatter,
			scatter src=explicit symbolic, %
			scatter/classes={
				REALIZABLE={draw=Dark2-C, mark=x},
				UNREALIZABLE={draw=Dark2-C, mark=x}
			}
		}
	}
	\begin{subfigure}{\textwidth}
		\begin{minipage}{0.32\textwidth}
			\begin{tikzpicture}
				\begin{axis}[plot,status colors,xmode=log,ymode=log,xlabel=\SemMLOne,ylabel=\SemML]
					\axislines{10000}{10}
					\addplot+[only marks,mark size=\plotmarksize]
					table [x=size_SemmlTac, y=size_Semml, meta=status_gt] {data/output_csvs/pairwise_aiger_size_Semml_SemmlTac.csv};
				\end{axis}
			\end{tikzpicture}
		\end{minipage}
		\begin{minipage}{0.32\textwidth}
			\begin{tikzpicture}
				\begin{axis}[plot,status colors,xmode=log,ymode=log,xlabel=\Strix]
					\axislines{10000}{10}
					\addplot+[only marks,mark size=\plotmarksize]
					table [x=size_Strix, y=size_Semml, meta=status_gt] {data/output_csvs/pairwise_aiger_size_Semml_Strix.csv};
				\end{axis}
			\end{tikzpicture}
		\end{minipage}
		\begin{minipage}{0.32\textwidth}
			\begin{tikzpicture}
				\begin{axis}[plot,status colors,xmode=log,ymode=log,xlabel=\ltlsynt]
					\axislines{10000}{10}
					\addplot+[only marks,mark size=\plotmarksize]
					table [x=size_Ltlsynt, y=size_Semml, meta=status_gt] {data/output_csvs/pairwise_aiger_size_Semml_Ltlsynt.csv};
				\end{axis}
			\end{tikzpicture}
		\end{minipage}
		\subcaption{Pairwise quality information of \runMealy{} track, measured by number of latches + number of gates.}
	\end{subfigure}
	
	\caption{Pairwise scatterplots to compare the solution sizes of two tools on a benchmark level.
		The axes are in log-log format, meaning that only slight deviations from the main diagonal already constitute a large difference.
		In particular, the dashed lines already mark a difference by a factor of \textbf{10}.}
		\label{fig:size_scatter}

\end{figure}
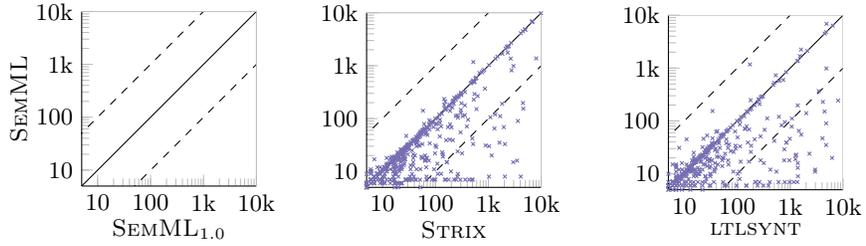
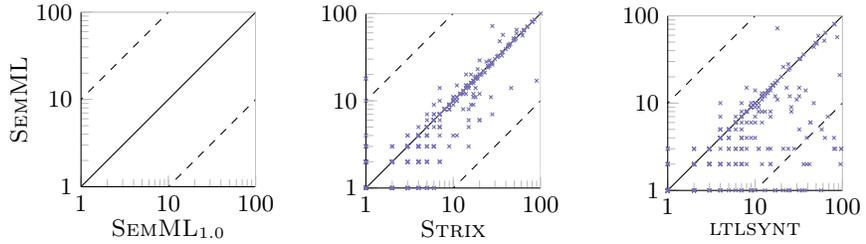
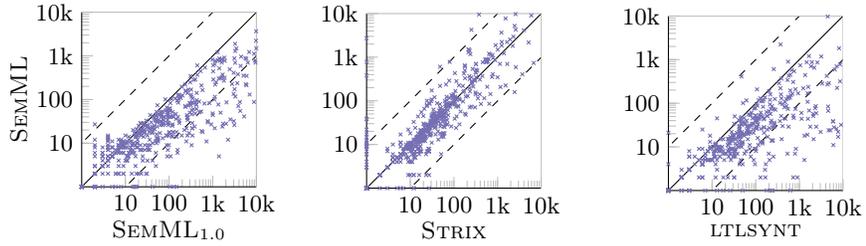

For the 2 synthesis tracks, we provide pairwise scatterplots to compare the solutions benchmark by benchmark (\cref{fig:size_scatter}).

\subsubsection{Impact of \SemML's Solution Extraction Pipeline.}
Our new solution pipeline not only heavily improves the time it takes to extract solutions but also their quality.
When comparing the \runAiger{} track between \SemML{} and our previous version, we see that most samples, heavily profit from out changes, with a notable amount being improved by over an order of magnitude.
Further, of course Mealy machines were not even possible in the prior version.

\subsubsection{Mealy Machines of \Strix{} and \LtlSynt{}.}
Comparing the Mealy machines of \SemML{} to \Strix{} and \LtlSynt{} (which both use the same (original) version of \textsf{MeMin}) we observe that we get significantly smaller machines.
To assess, how much of this difference is caused by our adaptation of \textsf{MeMin} being able to find more matching states, and how much is caused by it finding more matching edges (and especially not decomposing them), we also report the number of states per machine in \cref{subfig:states}.
We see that compared to both tools, our algorithm finds a notable amount of machines with fewer states, in the case of \ltlsynt{} even several, an order of magnitude smaller.
However, a good part of our observed reduction in size is due to our machines using fewer edges.
In parts, this is certainly due to fewer states, requireing fewer edges, and especially, our algorithm being able to cover more original edges with fewer representatives due to having more options.
However, we conjecture that the larger part of this difference stems from our algorithm not decomposing edges to have cube input labels.

\subsubsection{Aiger Circuits of \Strix.}
Out of the three tools, \Strix{} by far has the highest quality in Aiger solutions.
As highlighted in RQ3, we manage to match the quality of \Strix, while solving more instances and doing so faster.
Additionally, the scatterplot for \SemML{} and \Strix{} on the \runAiger{} track shows that there is quite a bit of variance.
In other words, there are several samples in which we produce much smaller circuits and vice versa (recall that the dashed lines mark a tenfold difference).
This shows great potential for both tools to appear in a portfolio approach such as \cite{DBLP:conf/tacas/CoslerHOS24}.

\begin{figure}[p]
	\setlength{\tabcolsep}{4pt}
	\renewcommand{\arraystretch}{1.5}
	\begin{subfigure}{\textwidth}
		\centering
		\subcaption{Confusion matricies for the \runReal{} track}
		\resizebox{0.9\textwidth}{!}{
		\begin{tabular}{ll @{\hspace{17pt}}c@{\hspace{6pt}}c c @{\hspace{17pt}}c@{\hspace{6pt}}c c @{\hspace{17pt}}c@{\hspace{6pt}}c}
			\toprule
			&& \multicolumn{2}{c}{\SemMLOne} && \multicolumn{2}{c}{\Strix} &&                     \multicolumn{2}{c}{\ltlsynt}                      \\
			\cmidrule(l{2pt}r{2pt}){3-4} \cmidrule(l{2pt}r{2pt}){6-7} \cmidrule(l{2pt}r{2pt}){9-10}
			&& solved & unsolved && solved & unsolved && solved & unsolved             \\ \midrule			\multirow{2}{*}{\rotatebox[origin=c]{90}{\SemML}} 
			&  solved  & \SemmlSemmlTacRealSolvesBoth  &  \SemmlSemmlTacRealUniqueSolvesTotal && \SemmlStrixRealSolvesBoth  &  \SemmlStrixRealUniqueSolvesTotal&&\SemmlLtlsyntRealSolvesBoth  &  \SemmlLtlsyntRealUniqueSolvesTotal \\
			& unsolved & \SemmlTacSemmlRealUniqueSolvesTotal & \SemmlSemmlTacRealSolvesNeither && \StrixSemmlRealUniqueSolvesTotal & \SemmlStrixRealSolvesNeither &&\LtlsyntSemmlRealUniqueSolvesTotal & \SemmlLtlsyntRealSolvesNeither \\
			\bottomrule
		\end{tabular}
	}
	\end{subfigure}
	\vspace{0.5cm}
	
	\begin{subfigure}{\textwidth}
		\centering
		\subcaption{Confusion matricies for the \runMealy{} track}
		\resizebox{0.9\textwidth}{!}{
			\begin{tabular}{ll @{\hspace{17pt}}c@{\hspace{6pt}}c c @{\hspace{17pt}}c@{\hspace{6pt}}c c @{\hspace{17pt}}c@{\hspace{6pt}}c}
				\toprule
				&& \multicolumn{2}{c}{\SemMLOne} && \multicolumn{2}{c}{\Strix} &&                     \multicolumn{2}{c}{\ltlsynt}                      \\
				\cmidrule(l{2pt}r{2pt}){3-4} \cmidrule(l{2pt}r{2pt}){6-7} \cmidrule(l{2pt}r{2pt}){9-10}
				&& solved & unsolved && solved & unsolved && solved & unsolved             \\ \midrule			\multirow{2}{*}{\rotatebox[origin=c]{90}{\SemML}} 
				&  solved  & -  &  - && \SemmlStrixMealySolvesBoth  &  \SemmlStrixMealyUniqueSolvesTotal&&\SemmlLtlsyntMealySolvesBoth  &  \SemmlLtlsyntMealyUniqueSolvesTotal \\
				& unsolved & - & - && \StrixSemmlMealyUniqueSolvesTotal & \SemmlStrixMealySolvesNeither &&\LtlsyntSemmlMealyUniqueSolvesTotal & \SemmlLtlsyntMealySolvesNeither \\
				\bottomrule
			\end{tabular}
		}
	\end{subfigure}
	\vspace{0.5cm}
	
	\begin{subfigure}{\textwidth}
		\centering
		\subcaption{Confusion matricies for the \runAiger{} track}
		\resizebox{0.9\textwidth}{!}{
		\begin{tabular}{ll @{\hspace{17pt}}c@{\hspace{6pt}}c c @{\hspace{17pt}}c@{\hspace{6pt}}c c @{\hspace{17pt}}c@{\hspace{6pt}}c}
			\toprule
			&& \multicolumn{2}{c}{\SemMLOne} && \multicolumn{2}{c}{\Strix} &&                     \multicolumn{2}{c}{\ltlsynt}                      \\
			\cmidrule(l{2pt}r{2pt}){3-4} \cmidrule(l{2pt}r{2pt}){6-7} \cmidrule(l{2pt}r{2pt}){9-10}
			&& solved & unsolved && solved & unsolved && solved & unsolved             \\ \midrule			\multirow{2}{*}{\rotatebox[origin=c]{90}{\SemML}} 
			&  solved  & \SemmlSemmlTacAigerSolvesBoth  &  \SemmlSemmlTacAigerUniqueSolvesTotal && \SemmlStrixAigerSolvesBoth  &  \SemmlStrixAigerUniqueSolvesTotal&&\SemmlLtlsyntAigerSolvesBoth  &  \SemmlLtlsyntAigerUniqueSolvesTotal \\
			& unsolved & \SemmlTacSemmlAigerUniqueSolvesTotal & \SemmlSemmlTacAigerSolvesNeither && \StrixSemmlAigerUniqueSolvesTotal & \SemmlStrixAigerSolvesNeither &&\LtlsyntSemmlAigerUniqueSolvesTotal & \SemmlLtlsyntAigerSolvesNeither \\
			\bottomrule
		\end{tabular}
		}
	\end{subfigure}

	\caption{Confusion matrices to highlight unique and overlapping solves for every pair of tools}
	\label{fig:confusion_matrices}

\end{figure}

\begin{figure}[p]
	\pgfplotsset{
		table/col sep=comma,
		plot/.style={
			width=\textwidth,height=\textwidth,
			x label style={anchor=north,inner sep=0pt},
			y label style={anchor=south,inner sep=0pt},
			xtick={2,60,1800}, xticklabels={2,60,1800}, 
			ytick={2,60,1800}, yticklabels={2,60,1800},
			xmin=1,ymin=1,xmax=2000,ymax=2000,
			axis x line*=bottom,
			axis y line*=left
		},
		status colors/.style={
			scatter,
			scatter src=explicit symbolic, %
			scatter/classes={
				REALIZABLE={draw=Dark2-C, mark=x},
				UNREALIZABLE={draw=Dark2-C, mark=x}
			}
		}
	}
	\begin{subfigure}{\textwidth}
		\begin{minipage}{0.32\textwidth}
			\begin{tikzpicture}
				\begin{axis}[plot,status colors,xmode=log,ymode=log,xlabel=\SemMLOne,ylabel=\SemML]
					\axislines{1800}{2}
					\addplot+[only marks,mark size=\plotmarksize]
					table [x=time_SemmlTac, y=time_Semml, meta=status_gt] {data/output_csvs/pairwise_realizability_time_Semml_SemmlTac.csv};
				\end{axis}
			\end{tikzpicture}
		\end{minipage}
		\begin{minipage}{0.32\textwidth}
			\begin{tikzpicture}
				\begin{axis}[plot,status colors,xmode=log,ymode=log,xlabel=\Strix]
					\axislines{1800}{2}
					\addplot+[only marks,mark size=\plotmarksize]
					table [x=time_Strix, y=time_Semml, meta=status_gt] {data/output_csvs/pairwise_realizability_time_Semml_Strix.csv};
				\end{axis}
			\end{tikzpicture}
		\end{minipage}
		\begin{minipage}{0.32\textwidth}
		\begin{tikzpicture}
			\begin{axis}[plot,status colors,xmode=log,ymode=log,xlabel=\ltlsynt]
				\axislines{1800}{2}
				\addplot+[only marks,mark size=\plotmarksize]
				table [x=time_Ltlsynt, y=time_Semml, meta=status_gt] {data/output_csvs/pairwise_realizability_time_Semml_Ltlsynt.csv};
			\end{axis}
		\end{tikzpicture}
		\end{minipage}
		\subcaption{Pairwise runtime information of \runReal{} track}
	\end{subfigure}
	
	\begin{subfigure}{\textwidth}
		\begin{minipage}{0.32\textwidth}
		
			\begin{tikzpicture}
				\begin{axis}[plot,status colors,xmode=log,ymode=log,xlabel=\SemMLOne,ylabel=\SemML]
					\axislines{1800}{2}
					\addplot+[only marks,mark size=\plotmarksize]
					table [x=time_SemmlTac, y=time_Semml, meta=status_gt] {data/helper_csvs/pairwise_mealy_time_Semml_SemmlTac.csv};
				\end{axis}
			\end{tikzpicture}
		
		\end{minipage}
		\begin{minipage}{0.32\textwidth}
			\begin{tikzpicture}
				\begin{axis}[plot,status colors,xmode=log,ymode=log,xlabel=\Strix]
					\axislines{1800}{2}
					\addplot+[only marks,mark size=\plotmarksize]
					table [x=time_Strix, y=time_Semml, meta=status_gt] {data/output_csvs/pairwise_mealy_time_Semml_Strix.csv};
				\end{axis}
			\end{tikzpicture}
		\end{minipage}
		\begin{minipage}{0.32\textwidth}
			\begin{tikzpicture}
				\begin{axis}[plot,status colors,xmode=log,ymode=log,xlabel=\ltlsynt]
					\axislines{1800}{2}
					\addplot+[only marks,mark size=\plotmarksize]
					table [x=time_Ltlsynt, y=time_Semml, meta=status_gt] {data/output_csvs/pairwise_mealy_time_Semml_Ltlsynt.csv};
				\end{axis}
			\end{tikzpicture}
		\end{minipage}
		\subcaption{Pairwise runtime information of \runMealy{} track}
	\end{subfigure}
	
	\begin{subfigure}{\textwidth}
		\begin{minipage}{0.32\textwidth}
			\begin{tikzpicture}
				\begin{axis}[plot,status colors,xmode=log,ymode=log,xlabel=\SemMLOne,ylabel=\SemML]
					\axislines{1800}{2}
					\addplot+[only marks,mark size=\plotmarksize]
					table [x=time_SemmlTac, y=time_Semml, meta=status_gt] {data/output_csvs/pairwise_aiger_time_Semml_SemmlTac.csv};
				\end{axis}
			\end{tikzpicture}
		\end{minipage}
		\begin{minipage}{0.32\textwidth}
			\begin{tikzpicture}
				\begin{axis}[plot,status colors,xmode=log,ymode=log,xlabel=\Strix]
					\axislines{1800}{2}
					\addplot+[only marks,mark size=\plotmarksize]
					table [x=time_Strix, y=time_Semml, meta=status_gt] {data/output_csvs/pairwise_aiger_time_Semml_Strix.csv};
				\end{axis}
			\end{tikzpicture}
		\end{minipage}
		\begin{minipage}{0.32\textwidth}
			\begin{tikzpicture}
				\begin{axis}[plot,status colors,xmode=log,ymode=log,xlabel=\ltlsynt]
					\axislines{1800}{2}
					\addplot+[only marks,mark size=\plotmarksize]
					table [x=time_Ltlsynt, y=time_Semml, meta=status_gt] {data/output_csvs/pairwise_aiger_time_Semml_Ltlsynt.csv};
				\end{axis}
			\end{tikzpicture}
		\end{minipage}
		\subcaption{Pairwise runtime information of \runAiger{} track}
	\end{subfigure}
	
	\caption{Pairwise scatterplots to compare the solve times of two tools on a benchmark level.
	The axes are in log-log format, meaning that only slight deviations from the main diagonal already constitute a large difference.
	In particular, the dashed lines already mark a difference by a factor of \textbf{2}.
	Note, that for the sake of readability, these plots start their axes at 1, cutting of some samples.
	The only notable samples that are missing are the \enquote{direct strategy} samples that \LtlSynt{} solves in under a second whereas \SemML{}, (as well as \Strix{}) timeouts.}
	\label{fig:time_scatter}

\end{figure}
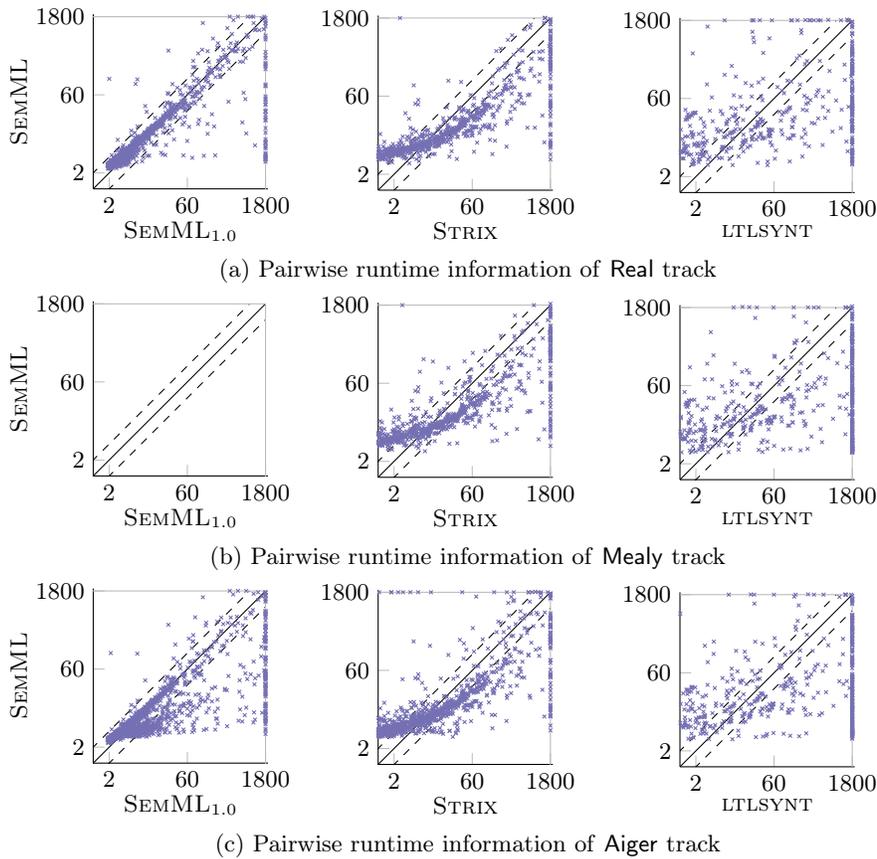

\begin{figure}[p]
	
	\begin{subfigure}{\textwidth}
	\centering
	\subcaption{Speedup factors for the \runReal{} track, for different lower cutoffs.}
	\resizebox{0.9\textwidth}{!}{
		\begin{tabular}{l@{\hspace{13pt}}c@{\hspace{13pt}}c@{\hspace{13pt}}c@{\hspace{13pt}}c@{\hspace{13pt}}c}
			\toprule
			& 0 & 5& 10& 30 & 120\\
			\midrule
			\SemMLOne & \SemmlSemmlTacRealTimeWtoFacZero{~\scriptsize(\SemmlSemmlTacRealTimeWtoMeanNZero)}&\SemmlSemmlTacRealTimeWtoFacFive{~\scriptsize(\SemmlSemmlTacRealTimeWtoMeanNFive)}&\SemmlSemmlTacRealTimeWtoFacTen{~\scriptsize(\SemmlSemmlTacRealTimeWtoMeanNTen)}&\SemmlSemmlTacRealTimeWtoFacThirty{~\scriptsize(\SemmlSemmlTacRealTimeWtoMeanNThirty)}&\SemmlSemmlTacRealTimeWtoFacOneTwenty{~\scriptsize(\SemmlSemmlTacRealTimeWtoMeanNOneTwenty)} \\
			\Strix    & \SemmlStrixRealTimeWtoFacZero{~\scriptsize(\SemmlStrixRealTimeWtoMeanNZero)}&\SemmlStrixRealTimeWtoFacFive{~\scriptsize(\SemmlStrixRealTimeWtoMeanNFive)}&\SemmlStrixRealTimeWtoFacTen{~\scriptsize(\SemmlStrixRealTimeWtoMeanNTen)}&\SemmlStrixRealTimeWtoFacThirty{~\scriptsize(\SemmlStrixRealTimeWtoMeanNThirty)}&\SemmlStrixRealTimeWtoFacOneTwenty{~\scriptsize(\SemmlStrixRealTimeWtoMeanNOneTwenty)} \\
			\LtlSynt  & \SemmlLtlsyntRealTimeWtoFacZero{~\scriptsize(\SemmlLtlsyntRealTimeWtoMeanNZero)}&\SemmlLtlsyntRealTimeWtoFacFive{~\scriptsize(\SemmlLtlsyntRealTimeWtoMeanNFive)}&\SemmlLtlsyntRealTimeWtoFacTen{~\scriptsize(\SemmlLtlsyntRealTimeWtoMeanNTen)}&\SemmlLtlsyntRealTimeWtoFacThirty{~\scriptsize(\SemmlLtlsyntRealTimeWtoMeanNThirty)}&\SemmlLtlsyntRealTimeWtoFacOneTwenty{~\scriptsize(\SemmlLtlsyntRealTimeWtoMeanNOneTwenty)}  \\
			\bottomrule
		\end{tabular}
	}
	\end{subfigure}
	\vspace{0.5cm}
	
	\begin{subfigure}{\textwidth}
		\centering
		\subcaption{Speedup factors for the \runMealy{} track, for different lower cutoffs.}
		\resizebox{0.9\textwidth}{!}{
			\begin{tabular}{l@{\hspace{13pt}}c@{\hspace{13pt}}c@{\hspace{13pt}}c@{\hspace{13pt}}c@{\hspace{13pt}}c}
				\toprule
				& 0 & 5& 10& 30 & 120\\
				\midrule
				\SemMLOne & -&-&-&-&-\\
				\Strix    & \SemmlStrixMealyTimeWtoFacZero{~\scriptsize(\SemmlStrixMealyTimeWtoMeanNZero)}&\SemmlStrixMealyTimeWtoFacFive{~\scriptsize(\SemmlStrixMealyTimeWtoMeanNFive)}&\SemmlStrixMealyTimeWtoFacTen{~\scriptsize(\SemmlStrixMealyTimeWtoMeanNTen)}&\SemmlStrixMealyTimeWtoFacThirty{~\scriptsize(\SemmlStrixMealyTimeWtoMeanNThirty)}&\SemmlStrixMealyTimeWtoFacOneTwenty{~\scriptsize(\SemmlStrixMealyTimeWtoMeanNOneTwenty)} \\
				\LtlSynt  & \SemmlLtlsyntMealyTimeWtoFacZero{~\scriptsize(\SemmlLtlsyntMealyTimeWtoMeanNZero)}&\SemmlLtlsyntMealyTimeWtoFacFive{~\scriptsize(\SemmlLtlsyntMealyTimeWtoMeanNFive)}&\SemmlLtlsyntMealyTimeWtoFacTen{~\scriptsize(\SemmlLtlsyntMealyTimeWtoMeanNTen)}&\SemmlLtlsyntMealyTimeWtoFacThirty{~\scriptsize(\SemmlLtlsyntMealyTimeWtoMeanNThirty)}&\SemmlLtlsyntMealyTimeWtoFacOneTwenty{~\scriptsize(\SemmlLtlsyntMealyTimeWtoMeanNOneTwenty)}  \\
				\bottomrule
			\end{tabular}
		}
	\end{subfigure}
	\vspace{0.5cm}
	
	\begin{subfigure}{\textwidth}
		\centering
		\subcaption{Speedup factors for the \runAiger{} track, for different lower cutoffs.}
		\resizebox{0.9\textwidth}{!}{
			\begin{tabular}{l@{\hspace{13pt}}c@{\hspace{13pt}}c@{\hspace{13pt}}c@{\hspace{13pt}}c@{\hspace{13pt}}c}
				\toprule
				& 0 & 5& 10& 30 & 120\\
				\midrule
				\SemMLOne & \SemmlSemmlTacAigerTimeWtoFacZero{~\scriptsize(\SemmlSemmlTacAigerTimeWtoMeanNZero)}&\SemmlSemmlTacAigerTimeWtoFacFive{~\scriptsize(\SemmlSemmlTacAigerTimeWtoMeanNFive)}&\SemmlSemmlTacAigerTimeWtoFacTen{~\scriptsize(\SemmlSemmlTacAigerTimeWtoMeanNTen)}&\SemmlSemmlTacAigerTimeWtoFacThirty{~\scriptsize(\SemmlSemmlTacAigerTimeWtoMeanNThirty)}&\SemmlSemmlTacAigerTimeWtoFacOneTwenty{~\scriptsize(\SemmlSemmlTacAigerTimeWtoMeanNOneTwenty)} \\
				\Strix    & \SemmlStrixAigerTimeWtoFacZero{~\scriptsize(\SemmlStrixAigerTimeWtoMeanNZero)}&\SemmlStrixAigerTimeWtoFacFive{~\scriptsize(\SemmlStrixAigerTimeWtoMeanNFive)}&\SemmlStrixAigerTimeWtoFacTen{~\scriptsize(\SemmlStrixAigerTimeWtoMeanNTen)}&\SemmlStrixAigerTimeWtoFacThirty{~\scriptsize(\SemmlStrixAigerTimeWtoMeanNThirty)}&\SemmlStrixAigerTimeWtoFacOneTwenty{~\scriptsize(\SemmlStrixAigerTimeWtoMeanNOneTwenty)} \\
				\LtlSynt  & \SemmlLtlsyntAigerTimeWtoFacZero{~\scriptsize(\SemmlLtlsyntAigerTimeWtoMeanNZero)}&\SemmlLtlsyntAigerTimeWtoFacFive{~\scriptsize(\SemmlLtlsyntAigerTimeWtoMeanNFive)}&\SemmlLtlsyntAigerTimeWtoFacTen{~\scriptsize(\SemmlLtlsyntAigerTimeWtoMeanNTen)}&\SemmlLtlsyntAigerTimeWtoFacThirty{~\scriptsize(\SemmlLtlsyntAigerTimeWtoMeanNThirty)}&\SemmlLtlsyntAigerTimeWtoFacOneTwenty{~\scriptsize(\SemmlLtlsyntAigerTimeWtoMeanNOneTwenty)}  \\
				\bottomrule
			\end{tabular}
		}
	\end{subfigure}
	\caption{
		Speedup factors for all tracks when taking different lower cutoffs.
		The number in a cell is calculated by taking the geometric mean of runtime ratios (\SemML{} over \textit{row-tool}) over all benchmarks where at least one tool required at least \textit{column} many seconds to solve.
		In particular, we cut out benchmarks where \textbf{both} tools required less than \textit{column} many seconds to focus on complex samples.
		The number in brackets denotes the number of remaining benchmarks used to calculate this geometric mean over, i.e.\ the number of benchmarks, (i) solved by at least one tool \textbf{and} (ii) at least one tool ran longer than the column's value in seconds.
		When a tool ran into the 30 minute timeout, we use that timeout value for the runtime ratio, which is an approximation that benefits tools with more timeouts
	}
	\label{fig:speedup_factors}
\end{figure}

